\documentclass[10pt,journal,compsoc,twocolumn ]{IEEEtran}

\usepackage{times}
\usepackage{epsfig}
\usepackage{graphicx}
\usepackage{amsmath}
\usepackage{amssymb}
\usepackage{stfloats}
\usepackage{multirow}
\usepackage{bbm}
\usepackage{rotating}
\usepackage{array}
\usepackage{color}
\usepackage{threeparttable}

\usepackage{epstopdf}
\usepackage{pifont}
\usepackage{subfigure}
\usepackage{bm}
\usepackage{url}
\usepackage{algorithm}
\usepackage{algorithmic}

\begin{document}

\title{Unsupervised Structure-Texture Separation Network for Oracle Character Recognition }

\author{Mei Wang, Weihong Deng, Cheng-Lin Liu
\thanks{Mei Wang and Weihong Deng are with the Pattern Recognition and Intelligent System Laboratory, School of Artificial Intelligence, Beijing University of Posts and Telecommunications, Beijing, 100876, China. E-mail: \{wangmei1,whdeng\}@bupt.edu.cn. (Corresponding Author: Weihong Deng)}
\thanks{Cheng-Lin Liu is with the National Laboratory of Pattern Recognition, Institute of Automation,
Chinese Academy of Sciences, Beijing, 100190, China, and also with the School of Artificial Intelligence, University of Chinese Academy of Sciences, Beijing, 100049, China. E-mail: liucl@nlpr.ia.ac.cn.}}

\maketitle

\begin{abstract}

Oracle bone script is the earliest-known Chinese writing system of the Shang dynasty and is precious to archeology and philology. However, real-world scanned oracle data are rare and few experts are available for annotation which make the automatic recognition of scanned oracle characters become a challenging task. Therefore, we aim to explore unsupervised domain adaptation to transfer knowledge from handprinted oracle data, which are easy to acquire, to scanned domain. We propose a structure-texture separation network (STSN), which is an end-to-end learning framework for joint disentanglement, transformation, adaptation and recognition. First, STSN disentangles features into structure (glyph) and texture (noise) components by generative models, and then aligns handprinted and scanned data in structure feature space such that the negative influence caused by serious noises can be avoided when adapting. Second, transformation is achieved via swapping the learned textures across domains and a classifier for final classification is trained to predict the labels of the transformed scanned characters. This not only guarantees the absolute separation, but also enhances the discriminative ability of the learned features. Extensive experiments on Oracle-241 dataset show that STSN outperforms other adaptation methods and successfully improves recognition performance on scanned data even when they are contaminated by long burial and careless excavation.

\end{abstract}


\section{Introduction}

Historical hieroglyphs are the major sources for knowing ancient civilization and have played an important role in archeology and philology. As the oldest hieroglyphs in China, oracle characters \cite{flad2008divination,keightley1997graphs} were inscribed on cattle bones or turtle shells about 3000 years ago and provided as key documentary records of human activities during that period. Therefore, the study of oracle characters is of vital importance for Chinese etymology and calligraphy as well as learning the culture and history of ancient China and even the world.

\begin{figure}
\centering
\includegraphics[width=8.8cm]{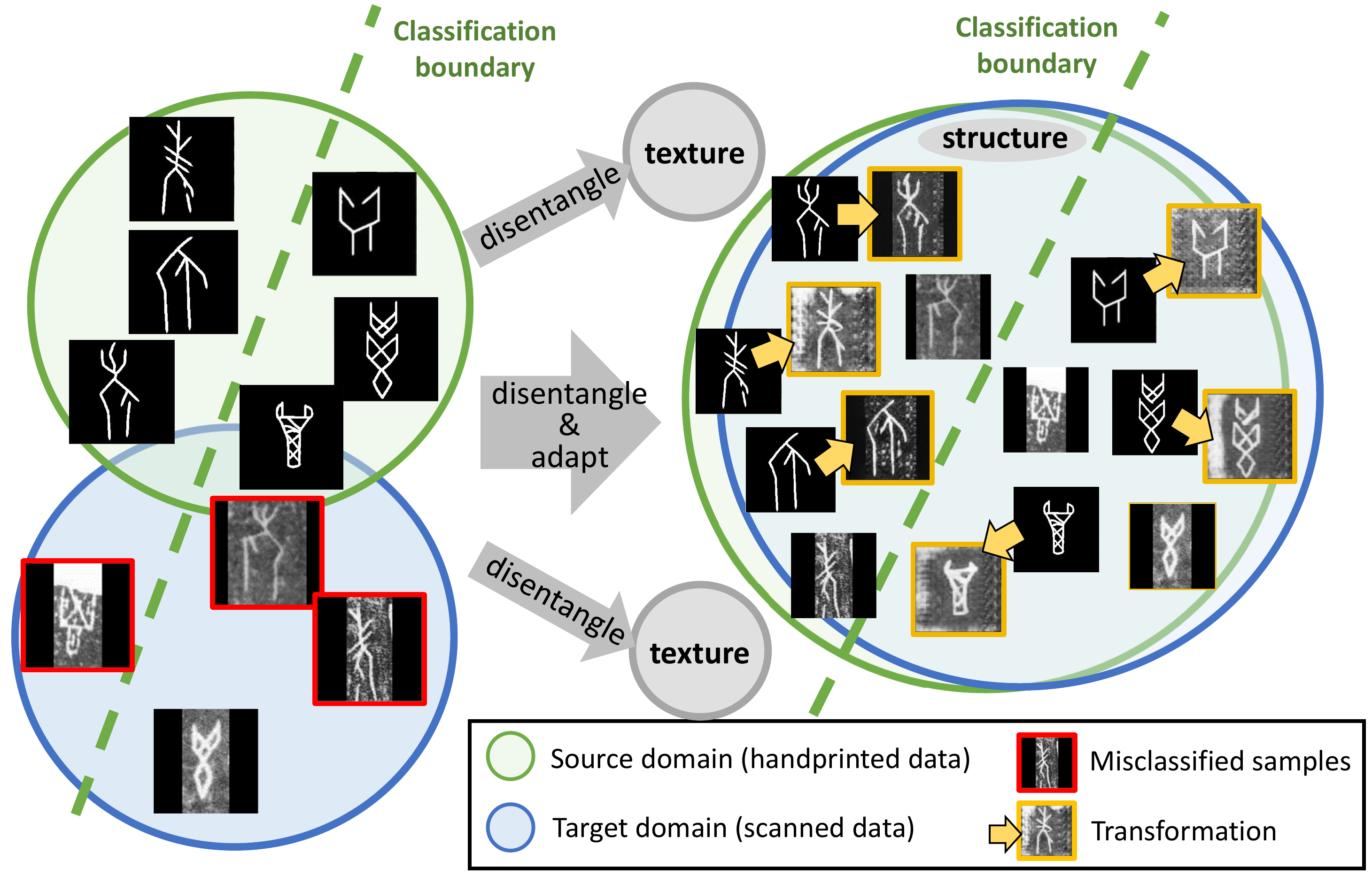}
\caption{The model trained on handprinted data cannot perform well on real scanned data due to distribution discrepancy (left). Benefiting from disentanglement, transformation and alignment, handprinted and scanned data are aligned in structure-shared space and the network is optimized by transformed characters, which improve performance on scanned data (right).}
\label{simple_arch} 
\end{figure}

Research on oracle characters began in the late of 1890s shortly after oracle bones were unearthed. Thus far, nearly 4,500 different oracle characters have been discovered, and only about 2,200 characters have been successfully deciphered \cite{li2020hwobc,huang2019obc306}. To help with the excavation of new oracle bones and the identification of unseen characters, deep convolutional neural networks (CNN) \cite{krizhevsky2012imagenet,simonyan2014very,szegedy2015going,he2016deep} are recently introduced to oracle character recognition \cite{guo2015building, huang2019obc306,zhang2019oracle}. However, training deep models always requires a large number of labeled samples, which becomes the major difficulty. 
On the one hand, oracle bones are extremely rare, and only limited number of scanned oracle data are available for training. On the other hand, 
scanned characters are broken and contain serious noises due to long burial and careless excavation, as shown in Fig. \ref{simple_arch} and \ref{oracle_example}. Annotating these data requires a high level of expertise and is extremely hard and time-consuming even for experts in archeology or paleography. Therefore, lacking of enough scanned data with labels is the main obstacle for the development of oracle character recognition systems.

In this paper, we propose to transfer knowledge acquired for handprinted character recognition to solve the recognition task of real scanned data. Handprinted oracle characters written by experts are available in large quantities, and can overcome the annotation burden. 
However, we experimentally prove that these high-quality data suffer from the distribution difference from real scanned data, which results in a dramatic performance drop when directly applying the model trained on handprinted data to recognize the newly excavated oracle characters, as shown in Fig. \ref{simple_arch}. 
Therefore, we propose to adopt unsupervised domain adaptation (UDA) \cite{wang2018deep} to mitigate the discrepancy, where labeled handprinted samples are set as source data and unlabeled scanned samples as target data. It aims to align the distributions between these two domains by finding a domain-invariant feature space \cite{Ganin2015Unsupervised,Long2015Learning,Long2016Unsupervised}, such that the learned model can generalize to scanned domain. While this, in principle, is a good idea, it still faces problems when adapting. 
1) Large domain shift between handprinted and scanned oracle data caused by serious noises makes it hard to align the two domains well, and this direct alignment leaves the learned domain-invariant representations vulnerable to the contamination with aspects of noises. 2) Different writing styles lead to high degrees of intra-class variance. Characters belonging to the same category largely vary in stroke and even topology which brings great difficulty for learning discriminative target features when lack of labels. 

To address this issue, we propose a structure-texture separation network (STSN) as shown in Fig. \ref{simple_arch}. First, our STSN learns a triplet of encoders to disentangle the features of individual domains into structure (e.g., glyph) and texture (e.g., noise and resolution) components, and then distribution alignment is safely taken in structure-related feature space. Unlike previous approaches \cite{Ganin2015Unsupervised,Long2015Learning} which directly align the entire domains in the whole feature space, separating and adapting the feature space in this manner can mitigate the negative influence caused by serious noises and abrasions. Second, inspired by the generative adversarial nets (GANs) \cite{goodfellow2014generative}, a generator and duplex discriminators are designed to transform images by swapping the learned texture information between any pairs of images, which guarantees the absolute separation of features via a min-max game rather than feature orthogonality \cite{bousmalis2016domain}. Third, benefitting from image transformation, data augmentation and label transfer are achieved at the same time for target domain. The transformed target-like data follow the distribution of scanned domain and maintain the category information transferred from handprinted domain, which addresses the problem of lacking target labels. A classifier is stacked on the structure representations for the final classification, which also predicts the labels of these target-like data, to pull the samples belonging to the same classes together and thus make target representations more generalized and discriminative.

To facilitate the research towards UDA for oracle character recognition, we present an \textit{Oracle-241} dataset\footnote{Data and code are available on https://github.com/wm-bupt/STSN}. It contains about \textit{80K images} of corresponding handprinted and scanned characters belonging to \textit{241 categories}. These images were selected from the oracle datasets collected by AnYang Normal University \cite{anyang}, and were rebuilt under unsupervised setting. Each image is centered by one oracle character. Oracle-241 contains extremely serious and unique noises caused by thousands of years of burial and aging, and contains various writing styles in most categories which all make it more difficult to recognize and adapt. Compared with the commonly-used UDA datasets, the categories of Oracle-241 are more than 20 times as many as those of MNIST-USPS-SVHN digit datasets \cite{lecun1998gradient,netzer2011reading,denker1989neural}.

Our contributions can be summarized into three aspects.

1) To the best of our knowledge, it is the first time that unsupervised domain adaptation is applied in oracle character recognition problem. This research aims to make recognition systems more practical and accurate when utilized to recognize real-world oracle characters, which contributes not only to technology but also to culture heritage preservation and the understanding of oracle characters and ancient civilization.

2) Our proposed STSN simultaneously learns to disentangle, transform, adapt and recognize. Separating texture from structure can avoid the negative influence caused by serious noises when adapting. Instead of transforming conditioned on a binary domain label, our STSN swaps the learned texture information between any pairs of images so that texture and structure of our transformed images are both from real images leading to realistic and diverse results. Simultaneously, the labels of the transformed images are further used in classifier training stage to improve the discrimination ability.

3) Extensive experimental results on a large Oracle-241 dataset show that our method successfully transfers recognition knowledge from handprinted oracle characters to scanned data. It outperforms other UDA methods and offers a state-of-the-art (SOTA) in performance on Oracle-241. Experiments on MNIST-USPS-SVHN digit datasets also show the effectiveness of the proposed method and prove that our STSN has the potential to generalize to a variety of scenarios.

\section{Related work}

\subsection{Oracle character recognition}

Identifying and deciphering oracle characters is one of the most important topics in oracle bone study. As technologies evolve, a handful of works have began to concentrate on oracle character recognition from the perspective of computer vision. Earliest researches \cite{zhou1995method,li2000coding,li2011recognition,shaotong2016identification} regarded oracle character as an undirected graph, and utilized its topological properties as features to  perform classification. 
Liu et al. \cite{liu2017oracle} extracted block histogram-based features and applied support vector machine (SVM) to recognize characters. Guo et al. \cite{guo2015building} collected a Oracle-20K dataset which composes of 20K handprinted oracle characters belonging to 261 categories and constructed a novel hierarchical representation. Recently, CNNs have achieved great progress in some computer vision tasks and are introduced into oracle character recognition. Huang et al. \cite{huang2019obc306} released a dataset of scanned oracle characters called OBC306 and presented the CNN-based evaluation for this dataset to serve as a benchmark. OracleNet \cite{lu2020recognition} considered the radical-level composition of oracle characters, and detected radicals using a Capsule network \cite{sabour2017dynamic} to recognize characters. Zhang et al. \cite{zhang2019oracle} trained a DenseNet \cite{huang2017densely} with triplet loss and performed classification by nearest neighbor classifier. Cross-modal oracle character recognition was first proposed by Zhang et al. \cite{zhang2021oracle}. They utilized adversarial learning and cross-modal triplet loss to perform classification on handprinted and scanned data in a supervised manner. However, our method focuses on UDA where the labels of scanned data are unavailable which is a more realistic but difficult scenario.

\subsection{Unsupervised domain adaptation}

Unsupervised domain adaptation \cite{wang2018deep} has been proposed to address domain shift between labeled source domain and unlabeled target domain. Popular UDA methods \cite{Long2015Learning,sun2016deep,zellinger2017central,peng2019moment} align distributions by moment matching. For example, maximum mean discrepancy (MMD) \cite{Long2015Learning,long2017deep,Yan2017Mind,chen2019graph} was applied to deep networks to reduce the distribution mismatch. 
Also, adversarial learning \cite{Tzeng2017Adversarial,Ganin2015Unsupervised,pei2018multi,chadha2019improved} has been widely used for alignment. Domain-adversarial neural network (DANN) \cite{Ganin2015Unsupervised} made domain classifier fail to predict the domain labels of features by a gradient reversal layer (GRL) such that the feature distributions over the two domains are similar. Conditional adversarial domain adaptation (CDAN) \cite{long2018conditional} conditioned an adversarial adaptation model on discriminative information conveyed in the classifier predictions. 
By borrowing the idea of GANs \cite{goodfellow2014generative}, generative adversarial methods \cite{hoffman2018cycada,hu2018duplex,sankaranarayanan2018generate} transformed source images to target-like ones to improve target performance. For example, cycle-consistent adversarial domain adaptation (CyCADA)  \cite{hoffman2018cycada} utilized CycleGAN \cite{zhu2017unpaired} to perform pixel-level and feature-level adaptation; while DupGAN \cite{hu2018duplex}  proposed a generator with duplex discriminators to restrict the latent representation to be domain invariant, with its category information preserved. Other methods \cite{bousmalis2016domain,zhang2018collaborative} explored to learn the disentangled feature representations to perform better adaptation. For example, domain separation network (DSN) \cite{bousmalis2016domain} applied feature orthogonality and image reconstruction to disentangle features into domain-specific and domain-invariant components, and performed alignment in domain-invariant space. In this paper, it is the first time that unsupervised domain adaptation is applied in oracle character recognition problem. Different from DSN, our STSN applies image transformation to constrain the disentanglement process. Besides, our transformed images are generated by using the disentangled texture information instead of the binary domain labels, and the transformed images are further utilized to optimize network to improve the performance on scanned domain in an unsupervised manner. 

\subsection{Text image editing}

Taken a content image and a style image as inputs, text image editing (TIE) \cite{wu2019editing,yang2020swaptext,shimoda2021rendering} aims to replace the text instance in the style image while retaining the styles of both the background and text. SRNet \cite{wu2019editing} first generated the foreground text by style transfer, then obtained the background image by text erasure, and finally synthesized the edited text image by fusing foreground and background. SwapText \cite{yang2020swaptext} further improved SRNet by manipulating geometric points of characters to transform text locations. Shimoda et al. \cite{shimoda2021rendering} formulated raster text editing as a de-rendering problem. They proposed a vectorization model to parse text information, and edited texts by a rendering engine using the parsed parameters. In principle, TIE can be utilized to mitigate the domain gap between handprinted and scanned images. Replacing the characters in scanned images with handprinted characters can generate labeled training images in scanned domain. However, there are several challenges when training such editing model for oracle characters. 1) Existing TIE methods require pairwise samples for training, i.e., a pair of images with same style but different texts. Although overlaying synthetic text to existing background images can generate pairwise synthetic images for TIE, it is quite difficult to apply this technology to oracle-bone images since background texture information is rare. 2) The ground-truths of style-transfer text and background image are also needed in TIE. However, the backgrounds and foregrounds of real-world scanned oracle images are difficult to segment due to serious noises. 3) Since oracle characters are less investigated and have not been digitalized, it is impossible to parse their rendering parameters, e.g., text, size and font. Considering of these challenges, we instead utilize image transformation in our STSN.

\section{Structure-texture separation network}

In this section, we describe our STSN method which simultaneously learns to disentangle, transform, adapt and recognize. By disentangling features, it not only aligns handprinted and scanned data in structure-shared feature space, but also swaps textures between any pairs of images to perform transformation. A classifier for final classification learns to predict the labels of transformed images which further enhances the discriminative ability of the learned features.

\subsection{Overview}

In our study, we assume that there are $M$ labeled source domain (handprinted) samples $X^s=\{{x^{s}_{i}}\}^{M}_{i=1}$ with the corresponding labels $Y^s=\{{y^{s}_{i}}\}^{M}_{i=1}$, and $N$ unlabeled target domain (scanned) samples $X^t=\{{x^{t}_{i}}\}^{N}_{i=1}$ without any available annotated labels.
There is a discrepancy between the distributions of source and target domains. Our goal is to train a handprinted oracle character recognition system that can generalize well to scanned data through training on labeled source domain and unlabeled target domain. Specifically, we desire to obtain domain-invariant features that work equally well in both domains.

\begin{figure*}
\centering
\includegraphics[width=16cm]{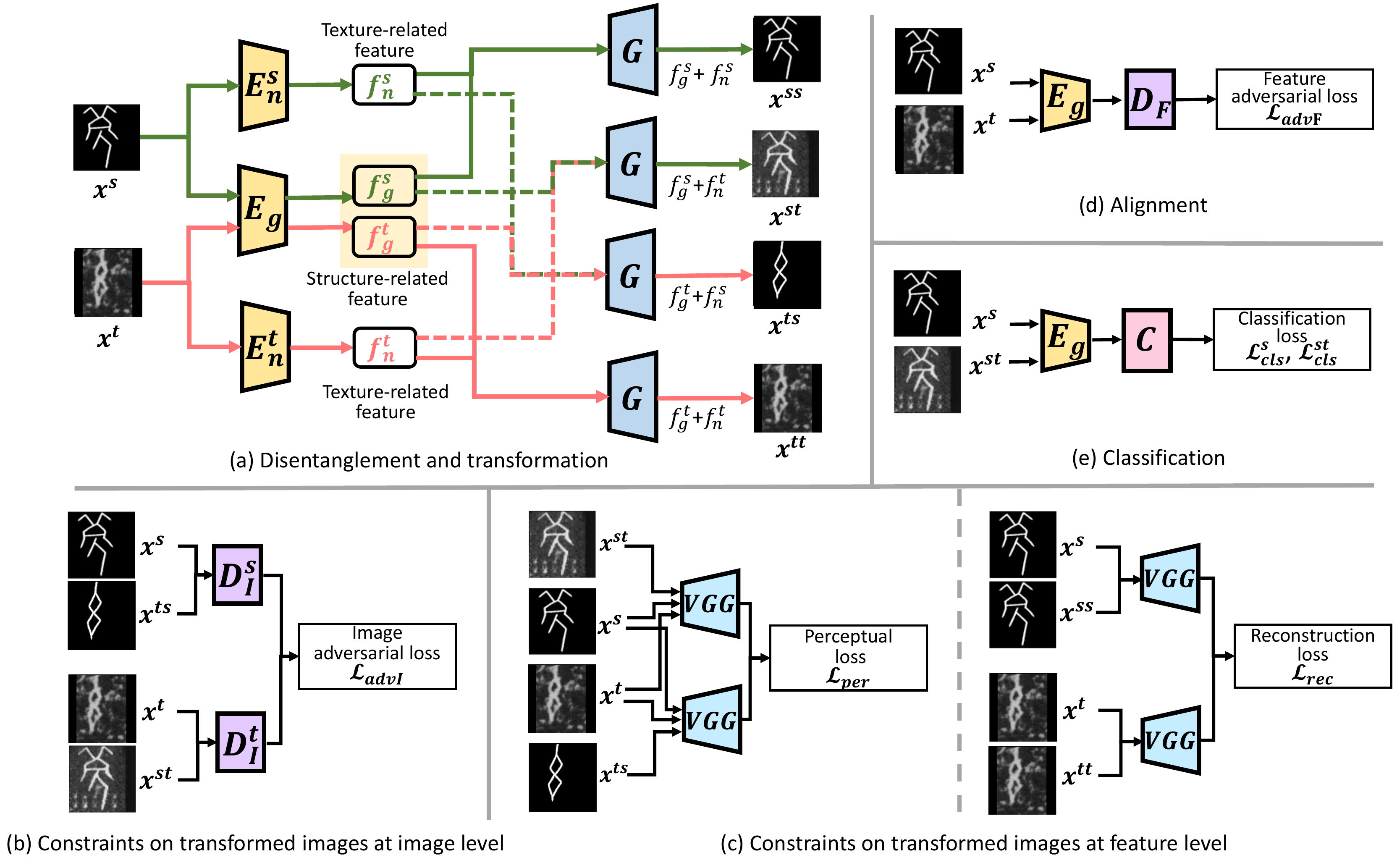}
\caption{Illustration of STSN. (a) $E_g$, $E_n^s$ and $E_n^t$ encode images into structure and texture feature space. Based on the disentangled features, $G$ generates the reconstructed and transformed images. (b) $D_I^s$ and $D_I^t$ are utilized to make the transformed images look real from image level. (c) VGGNet ensures the generated images contain proper structures and textures at feature level. (d) $D_F$ is pitted against $E_g$ to make structure-related features domain-invariant. (e) $C$ stacked on $E_g$ is trained on the source and transformed target-like images which further improves the generalization and discrimination of network. }
\label{architecture} 
\end{figure*}

An overview of the proposed method is shown in Fig. \ref{architecture}. The networks consist of a structure-shared encoder $E_g$, two texture-specific encoders $\{E_n^s, E_n^t\}$, a generator $G$, a feature-level discriminator $D_F$, two image-level discriminators $\{D_I^s, D_I^t\}$, a pretrained VGGNet, and a classifier $C$. Given $x^s$ and $x^t$,
\begin{itemize}
  \item \textbf{Encoder \bm{$E_g$}, \bm{$E_n^s$} and \bm{$E_n^t$}} disentangle and extract structure-shared and texture-specific features of the input images so that the contamination by texture can be avoided in adaptation process, as shown in Fig. \ref{architecture}(a).
  \item \textbf{Generator} \bm{$G$} utilizes the disentangled features to reconstruct the original input images and transform images across domains shown in Fig. \ref{architecture}(a). Better reconstruction and transformation would force encoders to disentangle well.
  \item \textbf{Discriminator \bm{$D_I^s$} and \bm{$D_I^t$}} are pitted against $G$ to ensure the transformed images look real from \emph{image level}. $D_I^s$ and $D_I^t$ distinguish between real and transformed images, while $G$ aims for fooling $D_I^s$ and $D_I^t$. They are optimized by image-level adversarial loss $\mathcal{L}_{advI}$ in Fig. \ref{architecture}(b).
  \item \textbf{VGGNet} aims to ensure the reconstructed and transformed images look real from \emph{feature level}. It extracts their perceptual features and constrains the feature similarity of structure and texture between the generated and real images supervised by reconstruction loss $\mathcal{L}_{rec}$ and perceptual loss $\mathcal{L}_{per}$, as shown in Fig. \ref{architecture}(c). 
  \item \textbf{Discriminator} \bm{$D_F$} is pitted against $E_g$ to align source and target features in structure-related space. 
  $D_F$ predicts the domain labels of features, while $E_g$ aims to fool $D_F$. They are guided by feature-level adversarial loss $\mathcal{L}_{advF}$ shown in Fig. \ref{architecture}(d).
  \item \textbf{Classifier} \bm{$C$} stacked on structure features tries to predict the labels of source and transformed target-like images supervised by classification loss $\mathcal{L}_{cls}^{s}$ and $\mathcal{L}_{cls}^{st}$, and thus reduces intra-class variance and improves the discriminative ability for target images, as shown in Fig. \ref{architecture}(e).
\end{itemize}

\subsection{Disentanglement and transformation} \label{encoder and generator}
\textbf{Disentanglement:} $E_g$ parameterized by $\theta_{E_g}$ aims for mapping an image $x$ from either source or target domain to a structure-shared representation $f_g$ that encodes high-level glyph information and is expected to be domain invariant. $E_n^s$ and $E_n^t$ parameterized by $\theta_{E_n^s}$ and $\theta_{E_n^t}$ map source image $x^s$ and target image $x^t$ to texture-specific representations $f_n^s$ and $f_n^t$ respectively, which encode low-level noise information and are private to each domain.
\begin{equation}
f_g= E_g\left (x\right ), x\in X^s\cup X^t,
\end{equation}
\begin{equation}
f^s_n = E^s_n\left (x^s\right ),
\end{equation}
\begin{equation}
f^t_n = E^t_n\left (x^t\right ).
\end{equation}

\textbf{Transformation:} To induce the model to produce such seprated representations and guarantee that $\{f^*_g,f^*_n\}$ are complementary to each other, image reconstruction and transformation are performed later. Specifically, the reconstructed image is generated from $G$ through recombining $f^*_g$ and $f^*_n$ of the same image:
\begin{equation}
x^{ss}=G\left ( z^{ss} \right )=G\left ( f_g^s,f_n^s,s \right ),
\end{equation}
\begin{equation}
x^{tt}=G\left ( z^{tt} \right )=G\left ( f_g^t,f_n^t,t \right ), \label{decoder_reconstruct}
\end{equation}
where the domain code $\left \{ s,t \right \}$ is used to specify which domain the latent representation is transformed to. For easy implementation, structure-shared feature, texture-specific feature and domain code are concatenated and utilized as the input of $G$, e.g., $z^{ss}=[f_g^s; f_n^s; s]$. We assume that the generated images $x^{ss}$ and $x^{tt}$ should be the same as the original input images, i.e., $x^{ss}\sim x^s$ and $x^{tt}\sim x^t$, which is constrained by reconstruction loss introduced in Section \ref{training loss}. Moreover, to transform images across domains, we swap the texture-specific components between any pairs of source and target images, and then decode them into two unseen images by $G$:
\begin{equation}
x^{st}=G\left ( z^{st} \right )=G\left ( f_g^s,f_n^t,t \right ),
\end{equation}
\begin{equation}
x^{ts}=G\left ( z^{ts} \right )=G\left ( f_g^t,f_n^s,s \right ), \label{decoder_transform}
\end{equation}
where $x^{st}$ is the transformed target-like image, which is generated by combining the structure of $x^s$ and the texture of $x^t$, and $x^{ts}$ is the transformed source-like image generated by the similar way. Intuitively, if $E_g$ and $E_n$ successfully encode the structure and texture representations respectively and these representations are complementary to each other as we expect, $x^{st}$ should look alike the real scanned image $x^t$ in terms of texture appearance while retaining the same glyph information as $x^s$. It is constrained by the image-level discriminator $D^t_I$ and perceptual loss introduced in Section \ref{training loss}. Similar constrains are applied on $x^{ts}$.

\subsection{Objective functions} \label{training loss}

The overall loss $\mathcal{L}$ of our framework consists of classification loss $\mathcal{L}_{cls}$, feature-level adversarial loss $\mathcal{L}_{advF}$, image-level adversarial loss $\mathcal{L}_{advI}$, perceptual loss $\mathcal{L}_{per}$, and reconstruction loss $\mathcal{L}_{rec}$. Our overall loss can be computed as follows,
\begin{equation}
\mathcal{L} = \mathcal{L}_{cls}+\alpha_1\mathcal{L}_{advF}+\alpha_2\mathcal{L}_{advI}+\alpha_3\mathcal{L}_{per}+\alpha_4\mathcal{L}_{rec}, \label{total}
\end{equation}
where $\{\alpha_i\}_{i=1}^4$ are the parameters for the trade-off between different terms. The detailed optimization procedure of our proposed STSN is depicted in the Algorithm \ref{al1}. The followings are the details of each loss.

\textbf{Classification Loss} (Fig. \ref{architecture}(e)). With the labeled source data, we are able to learn a basic recognition model in a supervised way,
\begin{equation}
\mathcal{L}_{cls}^{s}=\underset{C,E_g }{min}\mathbb{E}_{\left ( x^s,y^s \right )\sim \left ( X^s,Y^s \right )}L_{ce}\left ( C\left ( E_g(x^s) \right ),y^s \right ), \label{src_cls}
\end{equation}
where $L_{ce}(\cdot)$ is the cross-entropy loss function. 
Considering of the lack of target labels, the transformed target-like image $x^{st}$ is able to be utilized as an alternative to supervise learning proceeds for target domain. $x^{st}$ holds the same glyph information with the labeled source data $x^s$, so its category is expected to be the same as $x^s$'s. 
Thus, we optimize the network by the target-like images $x^{st}$ supervised with the labels of $x^s$,
\begin{equation}
\mathcal{L}_{cls}^{st}=\underset{C,E_g }{min}\mathbb{E}_{\left ( z^{st},y^s \right )\sim \left ( Z^{st},Y^s \right )}L_{ce}\left ( C\left ( E_g\left ( G\left ( z^{st} \right ) \right ) \right ),y^s \right ). \label{st_cls}
\end{equation}
Moreover, through transferring glyph information from $x^s$ to $x^{st}$, the transformed images $x^{st}$ consist of various writing styles as well. With help of $\mathcal{L}_{cls}^{st}$, STSN can push these samples of the same class together  and reduce intra-class variances for target domain. 
Therefore, the classification loss can be computed by $\mathcal{L}_{cls} = \mathcal{L}_{cls}^s + \mathcal{L}_{cls}^{st}$.

\textbf{Feature-level adversarial loss} (Fig. \ref{architecture}(d)). 
We optimize $D_F$ and $E_g$ in a min-max manner where $D_F$ aims to classify whether a structure-related feature is drawn from the source or the target domain ($D_F$ should output 0 for source features and 1 for the target ones), and $E_g$ aims to fool $D_F$ through generating domain-invariant features which reverse the outputs of $D_F$. The feature-level adversarial loss can be formulated as follows,
\begin{equation}
\begin{split}
\mathcal{L}_{advF}=\underset{E_g}{min}\ \underset{D_F}{max} &\ \mathbb{E}_{x^t\sim X^t}\left [ \log\left ( D_F\left ( E_g\left ( x^t \right ) \right ) \right ) \right ]\\
&+\mathbb{E}_{x^s\sim X^s}\left [ \log\left (1- D_F\left ( E_g\left ( x^s \right ) \right ) \right ) \right ]. \label{advf}
\end{split}
\end{equation}
By aligning source and target features into a domain-invariant space, we enable our model to learn on source data while still generalizing to target data. Moreover, our feature-level adversarial loss takes place in structure-related feature space so that the negative influence caused by serious noise is avoided when adapting leading to a better generalization.

\textbf{Image-level adversarial loss} (Fig. \ref{architecture}(b)). 
We employ image-level adversarial loss $\mathcal{L}_{advI} $ to encourage $G$  to generate realistic transformed images. $\mathcal{L}_{advI} $ is applied in two mapping directions, and can be expressed as,
\begin{equation}
\begin{split}
\mathcal{L}^t_{advI}=\underset{E_g,E^t_n,G}{min}\ & \underset{D^t_I}{max}\ \mathbb{E}_{x^t\sim X^t}\left [ \log\left ( D^t_{I}\left ( x^t \right ) \right )  \right ]\\
&+\mathbb{E}_{z^{st}\sim Z^{st}}\left [ \log \left ( 1-D^t_I\left ( G\left (  z^{st}\right ) \right ) \right ) \right ], \label{advi-t}
\end{split}
\end{equation}
\begin{equation}
\begin{split}
\mathcal{L}^s_{advI}=\underset{E_g,E^s_n,G}{min}\ & \underset{D^s_I}{max}\ \mathbb{E}_{x^s\sim X^s}\left [ \log\left ( D^s_{I}\left ( x^s \right ) \right )  \right ]\\
&+\mathbb{E}_{z^{ts}\sim Z^{ts}}\left [ \log \left ( 1-D^s_I\left ( G\left (  z^{ts}\right ) \right ) \right ) \right ]. \label{advi-s}
\end{split}
\end{equation}
For the mapping function $G:s+t\rightarrow st$, $D^t_I$ tries to distinguish real target data $x^t$ from the transformed target-like data $x^{st}$, while $G$ tries to fool $D^t_I$. Since $z^{st}=[E_g(x^s); E_n^t(x^t); t]$ is utilized as the input of $G$, the quality of transformed image is also highly dependent on $E_g$ and $E_n^t$. 
Therefore, we also optimize $E_g$ and $E_n^t$ to confuse $D^t_I$ so that they can well encode the structure and texture features for realistic transformation. This optimization contributes not only to $G$, but also to feature disentanglement and so our adaptation process. 
For the mapping function $G:t+s\rightarrow ts$ and the discriminator $D^s_I$, we impose the similar adversarial loss $ \mathcal{L}_{advI}^{s}$ for the transformed source-like image $x^{ts}$.
As proved in \cite{mao2017least}, the optimization of the standard GAN in Eq. (\ref{advi-t}) and (\ref{advi-s}) suffers from instability and may lead to the vanishing gradients problem. To address this issue, we utilize LSGAN \cite{mao2017least} instead. Therefore, our image-level adversarial loss can be computed by $\mathcal{L}_{advI} = \mathcal{L}_{advI}^s + \mathcal{L}_{advI}^{t}$.

\begin{figure*}[h]
\centering
\includegraphics[width=18cm]{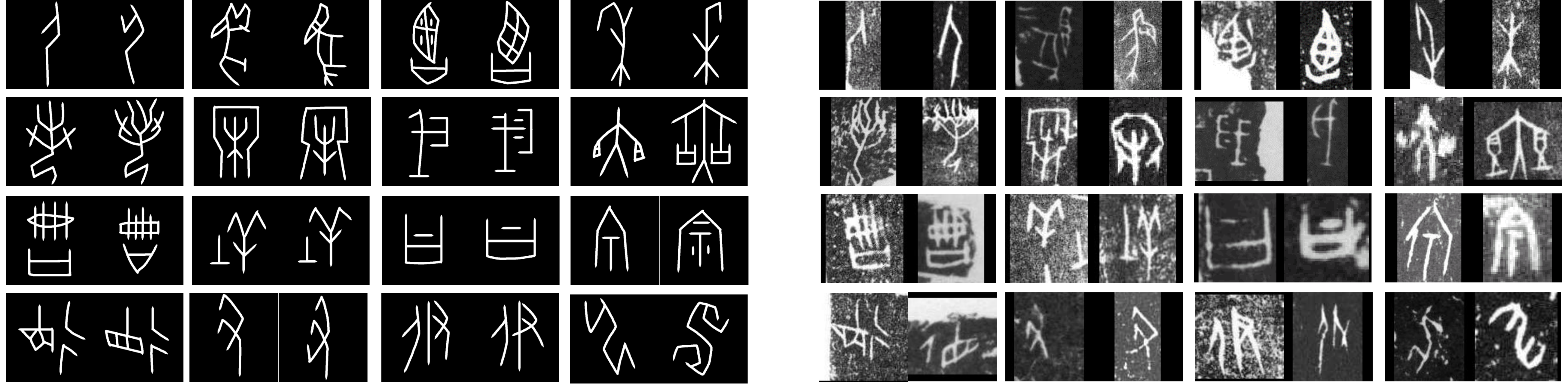}
\caption{The handprinted and scanned examples in Oracle-241 dataset. The left eight columns show handprinted samples belonging to 16 classes and the right eight columns show the corresponding scanned samples of the same 16 classes.}
\label{oracle_example} 
\end{figure*}

\textbf{Perceptual Loss} (Fig. \ref{architecture}(c)). $\mathcal{L}_{advI}$ only makes the transformed samples follow the distribution of real data without consideration of the glyph preservation when transforming. Therefore, we introduce the following perceptual loss to impose the constraints on both structure similarity and texture similarity. For the transformed image $x^{st}$, it constrains that $x^{st}$ is similar to $x^t$ in terms of texture because of the shared texture component $f^t_n$; it simultaneously requires that $x^{st}$ is similar to $x^s$ in terms of structure since they share the same $f^s_g$. Similar constraints are imposed on the transformed image $x^{ts}$. 
Some researches \cite{johnson2016perceptual,gatys2016neural} indicate that higher feature layers are good at keeping the structure information, while lower layers help to hold texture related things. Therefore, $\mathcal{L}_{per}$ is formulated as follows,
\begin{equation}
\begin{split}
\mathcal{L}_{per} = \underset{E_g,E^t_n,E^s_n,G}{min} &  \sum_{l\in \ell_{str}} \lambda_l (\left \| \phi_l\left ( x^s \right )- \phi_l\left ( G\left ( z^{st} \right ) \right )\right \|_1 \\
&  \ \ \ \ \ \ \ +\left \| \phi_l\left ( x^t \right )- \phi_l\left ( G\left ( z^{ts} \right ) \right )\right \|_1 ) \\
 + &\sum_{l\in \ell_{tex}}\lambda_l(  \left \| \mathcal{U}\left ( \phi_l\left ( x^t \right ) \right )- \mathcal{U}\left ( \phi_l\left ( G\left ( z^{st} \right ) \right ) \right )\right \|_1 \\
& \ \ \ \ \ \ \ + \left \| \mathcal{U}\left ( \phi_l\left ( x^s \right ) \right )- \mathcal{U}\left ( \phi_l\left ( G\left ( z^{ts} \right ) \right ) \right )\right \|_1) \label{per}
\end{split}
\end{equation}
where the first two terms represent structure loss which constrains the structure similarity; while the last two terms represent texture loss used to constrain the texture similarity. $\phi_l(\cdot)$ extracts the activation map from $l$-th layer of the pretrained VGGNet-16 model \cite{simonyan2014very}. The set of layers $\ell_{str}$ are the higher layers $\{\ell^{relu4\_1}, \ell^{relu5\_1}\}$ of VGGNet, while $\ell_{tex}$ are the lower layers $\{\ell^{relu1\_1}, \ell^{relu2\_1},\ell^{relu3\_1}\}$. $\lambda_l$ is trade-off parameter for $l$-th layer. Structure loss constrains L1 distance between the features extracted by higher layers; while texture loss constrains the channel-wisely feature similarity computed on lower layers inspired by AdaIN \cite{huang2017arbitrary}. In texture loss, $\mathcal{U}(\cdot)$ computes the mean activation across spatial dimensions independently for each channel,
\begin{equation}
 \mathcal{U}\left ( f \right )=  \frac{1}{HW}\sum_{h=1}^{H}\sum_{w=1}^{W}f_{chw},
\end{equation}
where 
$f_{chw} $ denotes $chw$-th element of the feature map, where $h$ and $w$ span spatial dimensions, $c$ is the feature channel. 

\textbf{Reconstruction loss} (Fig. \ref{architecture}(c)). As described previously in Section \ref{encoder and generator}, the reconstructed images $x^{ss}$ and $x^{tt}$ should be the same as the original input images $x^s$ and $x^t$. Therefore, we employ reconstruction loss to force them to have similar feature representations as computed by VGGNet,
\begin{equation}
\begin{split}
\mathcal{L}_{rec}=\underset{E_g,E^t_n,E^s_n, G}{min} \ \sum_{l\in \ell_{rec}} &\lambda_l \left \| \phi_l\left ( x^t \right )- \phi_l\left ( G\left ( z^{tt} \right ) \right )\right \|_1\\
&+\lambda_l \left \| \phi_l\left ( x^s \right )- \phi_l\left ( G\left ( z^{ss} \right ) \right )\right \|_1, \label{rec}
\end{split}
\end{equation}
where the set of layers $\ell_{rec}=\ell_{tex}\cup \ell_{str}$ consists of both lower and higher layers. 

\begin{algorithm}[htb]
\caption{ Optimization procedure of STSN.}
\label{al1}
\begin{algorithmic}[1]
\REQUIRE ~~\\
Labeled handprinted data $\{{x^{s}_{i}},{y^{s}_{i}}\}^{M}_{i=1}$, and unlabeled scanned data $\{{x^{t}_{i}}\}^{N}_{i=1}$.
\ENSURE ~~\\
The parameters of whole network $\Theta =\{\hat{\theta}_{E_g}$, $\hat{\theta}_{E_n^s}$, $\hat{\theta}_{E_n^t}$, $\hat{\theta}_{D_F}$, $\hat{\theta}_{D_I^s}$, $\hat{\theta}_{D_I^t}$, $\hat{\theta}_{G}$, $\hat{\theta}_C\}$.
\STATE Randomly initialize $\hat{\theta}_{E_n^s}$, $\hat{\theta}_{E_n^t}$, $\hat{\theta}_{D_F}$, $\hat{\theta}_{D_I^s}$, $\hat{\theta}_{D_I^t}$, $\hat{\theta}_{G}$ and $\hat{\theta}_C$, and initialize $\hat{\theta}_{E_g}$ with pretrained ImageNet model.
\REPEAT
\STATE Sample a minibatch of samples from both $X^s$ and $X^t$
\STATE / / Update discriminators by Eq. (\ref{advf}), (\ref{advi-t}) and (\ref{advi-s})
\STATE $\hat{\theta}_{D_F}=\theta_{D_F}-\eta_d \frac{\partial \mathcal{L}_{advF}}{\partial \theta_{D_F}}$
\STATE $\hat{\theta}_{D_I^t}=\theta_{D_I^t}-\eta_d \frac{\partial \mathcal{L}^{t}_{advI}}{\partial \theta_{D_I^t}}$, $\hat{\theta}_{D_I^s}=\theta_{D_I^s}-\eta_d \frac{\partial \mathcal{L}^{s}_{advI}}{\partial \theta_{D_I^s}}$
\STATE / / Update generator by Eq. (\ref{advi-t}), (\ref{advi-s}), (\ref{per}) and (\ref{rec})
\STATE $\hat{\theta}_{G}=\theta_{G}-\eta \alpha_2\frac{\partial \mathcal{L}_{advI} }{\partial \theta_{G}}-\eta \alpha_3\frac{\partial   \mathcal{L}_{per}}{\partial \theta_{G}}-\eta \alpha_4\frac{\partial\mathcal{L}_{rec} }{\partial \theta_{G}}$
\STATE / / Update encoders by Eq. (\ref{advf}), (\ref{advi-t}), (\ref{advi-s}), (\ref{per}) and (\ref{rec})
\STATE $\hat{\theta}_{E_n^t}=\theta_{E_n^t}-\eta\alpha_2\frac{\partial \mathcal{L}^{t}_{advI}}{\partial \theta_{E_n^t}}-\eta\alpha_3\frac{\partial\mathcal{L}_{per}}{\partial \theta_{E_n^t}}-\eta\alpha_4\frac{\partial\mathcal{L}_{rec} }{\partial \theta_{E_n^t}}$
\STATE $\hat{\theta}_{E_n^s}=\theta_{E_n^s}-\eta\alpha_2\frac{\partial \mathcal{L}^{s}_{advI}}{\partial \theta_{E_n^s}}-\eta\alpha_3\frac{\partial\mathcal{L}_{per}}{\partial \theta_{E_n^s}}-\eta\alpha_4\frac{\partial\mathcal{L}_{rec} }{\partial \theta_{E_n^s}}$
\STATE $\hat{\theta}_{E_g}=\theta_{E_g}-\eta_e\alpha_1 \frac{\partial \mathcal{L}_{advF} }{\partial \theta_{E_g}}-\eta_e\alpha_2\frac{\partial\mathcal{L}_{advI}}{\partial \theta_{E_g}}-\eta_e\alpha_3\frac{\partial\mathcal{L}_{per}}{\partial \theta_{E_g}}-\eta_e\alpha_4\frac{\partial\mathcal{L}_{rec}}{\partial \theta_{E_g}}-\eta_e \frac{\partial\mathcal{L}_{cls} }{\partial \theta_{E_g}}$
\STATE / / Update classifier by Eq. (\ref{src_cls}) and (\ref{st_cls})
\STATE $\hat{\theta}_{C}=\theta_{C}-\eta_e \frac{\partial \mathcal{L}_{cls}^{s}}{\partial \theta_{C}}-\eta_e \frac{\partial\mathcal{L}_{cls}^{st}  }{\partial \theta_{C}}$
\UNTIL{Convergence}
\end{algorithmic}
\end{algorithm}

\subsection{Discussion}

\textbf{Difference from DSN \cite{bousmalis2016domain}, GSN \cite{shi2018genre} and the work of Kim et al. \cite{kim2017adversarial}.} DSN, GSN and Kim et al. separated feature space by image reconstruction and feature orthogonality.  Orthogonality of structure and texture is sometimes too weak to guarantee the absolute separation. Moreover, intra-class variances caused by writing style are not able to be well addressed by only alignment. Image transformation is introduced into our STSN. Transformation contributes not only to feature disentanglement but also to data augmentation for target domain. 
Variances can be reduced by supervising learning process with the transformed target-like images.

\textbf{Difference from IDEL \cite{cheng2020improving}, DRLST \cite{john2019disentangled} and CAAE \cite{shen2017style}.} IDEL, DRLST and CAAE focus on text style transfer in the field of natural language processing. They learned the disentangled features, and aimed to generate style-transfer text. However, STSN focuses on a different recognition task, thus disentanglement and generation are just subtasks. Moreover, our method is different from them in terms of algorithm design. IDEL utilized mutual information to guarantee the disentanglement process, while STSN uses adversarial learning instead. DRLST only generated style-transfer text in inference stage, while STSN generates and utilizes the transformed images for network optimization in training stage. CAAE generated style-transfer samples conditioned on style labels, while STSN swaps the learned texture features to transform images. Instead of sharing the same encoder for learning the separated features, a triplet of encoders is utilized by STSN to avoid the negative mutual-effect.

\textbf{Difference from CyCADA \cite{hoffman2018cycada} and DupGAN \cite{hu2018duplex}.} All of CyCADA, DupGAN and our STSN follow the idea of GAN to achieve domain transformation for domain adaptation. CyCADA and DupGAN directly transformed images conditioned on a binary domain label. 
However, benefiting from feature disentanglement, the images in our STSN are generated by swapping texture-specific components between any pairs of source and target images so that texture and structure of our transformed images are both from real images.

\textbf{Difference from ASSDA \cite{zhang2021robust}, AHWR \cite{kang2020unsupervised} and the work of Luo et al. \cite{luo2021separating}.} ASSDA and AHWR focus on UDA of text recognition. Although they took the characteristics of text into consideration and proposed temporal pooling and character-level alignment to address the adaptation of sequence-like word images, they directly aligned domains in the whole feature space. In our STSN, structure information is disentangled from texture and alignment is adopted in structure-related feature space. Moreover, our method is capable of image transformation across domains, while ASSDA and AHWR are not. Luo et al. \cite{luo2021separating} aim to improve text recognition by removing backgrounds while retaining the text content. They shared attention masks from the recognizer to the discriminator to enable a character-level adversarial training when performing image transformation. However, our method is capable of feature disentanglement and alignment, while the work of Luo et al. is not.

\section{Experiments}

In this section, we perform UDA experiments on our Oracle-241 for oracle character recognition and on MNIST-USPS-SVHN \cite{lecun1998gradient,netzer2011reading,denker1989neural} for digit classification, which prove the effectiveness of our method. The details are as follows.

\subsection{Datasets}

\textbf{Oracle-241} is a large oracle character dataset for transferring knowledge from handprinted characters to scanned data, as shown in Fig. \ref{oracle_example}. Since oracle characters have not been digitalized, they are stored or displayed by images in computers. We selected the oracle images from the datasets collected by AnYang Normal University \cite{anyang} and built them under UDA setting. Our Oracle-241 contains \textit{78,565} handprinted and scanned \textit{characters} of \textit{241 categories}, and each image is centered by one single character. Scanned oracle images are generated by reproducing the oracle-bone surface by placing a piece of paper over the subject and then rubbing the paper with rolled ink. The aging process has made the inscriptions less legible so that these scanned characters are broken and exist serious noises. Handprinted oracle characters are written by experts, and their corresponding images are clean and high-definition. We split Oracle-241 into two subsets for training and testing as shown in Table \ref{tab4}. The training set consists of labeled 10,861 handprinted data and unlabeled 50,168 scanned data; while the testing set contains 3,730 handprinted data and 13,806 scanned data.

\begin{table}[htbp]
\renewcommand\arraystretch{1.1}
\caption{Statistic of our Oracle-241 dataset.}
    \label{tab4}
	\begin{center}
    \small
	\begin{tabular}{c|cc|cc}
    \hline
     \multirow{2}{*}{Subsets} & \multicolumn{2}{c|}{Train} & \multicolumn{2}{c}{Test}\\
        & \# Classes & \# Images & \# Classes   & \# Images \\ \hline \hline
        Handprint & 241 & 10,861 & 241 & 3,730 \\
        Scan & 241 &50,168 & 241 & 13,806 \\ \hline
	\end{tabular}
    \end{center}
\end{table}

\begin{table*}
\renewcommand\arraystretch{1.2}
\caption{Network Architectures of our Encoder and Generator used for oracle recognition. }
\label{archi_detail}
	\begin{center}
    \small
	\begin{tabular}{c|c}
		\hline
          Encoder ($E_n$) & Generator ($G$)  \\ \hline
          \textbf{Input:} $x\in \mathbb{R}^{224\times 224\times 3}$ & \textbf{Input:} $f_g, f_n,$ domain code  \\ \hline\hline
          Conv(k7n64s2), BN, ReLU & Deconv(k4n256s2), IN, ReLU, ConvBlock(k3n128s1)  \\
          Conv(k7n64s2), BN, ReLU & Deconv(k4n128s2), IN, ReLU, ConvBlock(k3n64s1)  \\
          Conv(k3n128s2), BN, ReLU & Deconv(k4n64s2), IN, ReLU, ConvBlock(k3n32s1)\\
          $\{$Conv(k3n256s2), BN, ReLU$\}\times$ 3 & $\{$Deconv(k4n32s2), IN, ReLU, ConvBlock(k3n32s1)$\}$$\times$ 2 \\\hline
          AvgPool, Conv(k1n8s1) & Conv(k1n3s1), Tanh \\ \hline\hline
          \textbf{Output: $f_n$ } & \textbf{Output:} $x^{ss}$/ $x^{st}$/ $x^{ts}$/ $x^{tt}$ \\ \hline
	\end{tabular}
    \end{center}
\end{table*}

\begin{table}
\renewcommand\arraystretch{1.2}
\caption{Network Architectures of our Discriminators used for oracle recognition. }
\label{archi_detail2}
	\begin{center}
    \small
	\begin{tabular}{c|c}
		\hline
          Discriminator ($D_I$) & Discriminator ($D_F$) \\ \hline
          \textbf{Input:} $x\in \mathbb{R}^{224\times 224\times 3}$ & \textbf{Input:} $f_g^s, f_g^t$ \\ \hline\hline
          Conv(k6n64s2), IN, LReLU(0.2) &Linear(1024), ReLU \\
          Conv(k4n128s2), IN, LReLU(0.2) & Dropout(0.5) \\
          Conv(k4n256s2), IN, LReLU(0.2) & Linear(1024), ReLU\\
           Conv(k4n512s2), IN, LReLU(0.2) & Dropout(0.5)\\\hline
           Linear(1) & Linear(1), Sigmoid \\ \hline\hline
           \textbf{Output:} Real/Fake & \textbf{Output:} Source/Target\\ \hline
	\end{tabular}
    \end{center}
\end{table}

\textbf{MNIST-USPS-SVHN \cite{lecun1998gradient,netzer2011reading,denker1989neural}} are the commonly used digit datasets for domain adaptation. They contain digital images of 10 classes. MNIST consists of a training set of 60,000 grey images and a testing set of 10,000 grey images. USPS contains 7,291 and 2,007 grey images for training and testing. SVHN contains 73,257 and 26,032 digits for training and testing. Following previous works \cite{hoffman2018cycada}, we perform domain adaptation on three transfer tasks: MNIST$\rightarrow$USPS (M$\rightarrow$U), USPS$\rightarrow$MNIST (U$\rightarrow$M) and SVHN$\rightarrow$MNIST (S$\rightarrow$M).

\subsection{Implementation detail}

\textbf{Network architecture.} For oracle character recognition, ResNet-18 \cite{he2016deep} pretrained on Imagenet \cite{russakovsky2015imagenet} is utilized as structure encoder $E_g$. Texture encoders $\{E_n^s, E_n^t\}$ contain 6 convolution blocks (Conv-BN-ReLU) followed by an average pooling layer and a 1$\times $1 convolution layer, and share the first convolution block with $E_g$. The texture information is encoded into a 8-dimension feature. For $G$ and $\{D_I^s,D_I^t\}$ networks, we use architectures similar to those used in DCGAN \cite{radford2015unsupervised}. We apply 5 stacked layer groups as $G$, and each layer group consists of a 4$\times $4 transposed convolution layer \cite{dumoulin2016guide}, one IN layer \cite{ulyanov2016instance}, one ReLU layer and one convolution block (Conv-LN-ReLU). Then 1$\times $1 convolution layer is applied to yield $224\times 224\times 3$ images. $\{D_I^s,D_I^t\}$ contain 4 convolution blocks (Conv-IN-LReLU), followed by a fully connected layer. For $D_F$, we use the same discriminative network structure as DANN \cite{Ganin2015Unsupervised}. Details are shown in Table \ref{archi_detail} and \ref{archi_detail2}.

For digit classification, we use a variant of the LeNet architecture as $E_g$ following recent works \cite{hoffman2018cycada,long2018conditional}. The same architecture is applied to $\{E_n^s, E_n^t\}$. Other networks are similar to those used in oracle character recognition.

\begin{table*}
\renewcommand\arraystretch{1.1}
\small
\caption{Ablation study on Oracle-241 dataset. }
\label{ablation}
	\begin{center}
    \setlength{\tabcolsep}{3.5mm}{
	\begin{tabular}{c|cccccc|c}
		\hline
          \multirow{2}{*}{Methods} & \multicolumn{6}{c|}{Components} &  \multirow{2}{*}{Scan} \\
         & $\mathcal{L}_{cls}^s$  & $\mathcal{L}_{advF}$ & $\mathcal{L}_{advI}$ & $\mathcal{L}_{rec}$ & $\mathcal{L}_{per}$ & $\mathcal{L}_{cls}^{st}$ &  \\ \hline \hline
         Source-only & \ding{51}& \textcolor{red}{\ding{55}} & \textcolor{red}{\ding{55}} & \textcolor{red}{\ding{55}}  & \textcolor{red}{\ding{55}} & \textcolor{red}{\ding{55}} &2.1   \\
         Model-A &\ding{51}& \textcolor{red}{\ding{55}} & \ding{51} &  \ding{51} & \ding{51} & \ding{51} & 31.8  \\
         Model-B &\ding{51}&  \ding{51} & \textcolor{red}{\ding{55}} & \textcolor{red}{\ding{55}} & \textcolor{red}{\ding{55}} & \textcolor{red}{\ding{55}} & 39.5   \\
         Model-C &\ding{51}& \ding{51}& \ding{51} & \textcolor{red}{\ding{55}}  & \textcolor{red}{\ding{55}} &  \textcolor{red}{\ding{55}} & 40.3  \\
         Model-D &\ding{51}& \ding{51} & \ding{51} &  \ding{51} &  \textcolor{red}{\ding{55}}&  \textcolor{red}{\ding{55}} &  41.2 \\
         Model-E &\ding{51}& \ding{51} & \ding{51} &  \ding{51} & \ding{51} & \textcolor{red}{\ding{55}} & 42.7  \\
         \textbf{STSN (ours)} &\ding{51}& \ding{51} & \ding{51} &  \ding{51} & \ding{51} & \ding{51} &\textbf{47.1}   \\\hline
	\end{tabular}}
    \end{center}
\end{table*}

\textbf{Experimental setup.} We train our model for 150,000 iterations over the data with batch size 16. For preprocessing, we randomly crop and flip the training samples. We adopt mini-batch SGD with momentum of 0.9 to train $E_g$ and $C$, and set weight decay and initial learning rate to 5$e-$4 and 2.5$e-$4, respectively. We use Adam solver with base learning rate of 1$e-$4 and momentum of 0.99 to train the discriminators $D_F$ and $\{D_I^s,D_I^t\}$, and use Adam solver with base learning rate of 1$e-$3 and momentum of 0.99 to train $\{E_n^s, E_n^t\}$ and $G$. The learning rate is annealed by $\eta=\eta_0\left ( \frac{1-T}{T_{max}} \right )^{0.9}$, where $\eta_0$ is the initial learning rate, $T$ and $T_{max}$ are the current and total iteration. The parameters $\{\alpha_i\}_{i=1}^4$ are set to be 1, 0.01, 0.05 and 0.5, respectively. $\{\lambda_l\}_{l\in \ell_{rec}}$ are set to be 1/32, 1/16, 1/8, 1/4 and 1, respectively, which emphasize more on higher layer. $\{\lambda_l\}_{l\in \ell_{str}}$ share the same value with $\{\lambda_l\}_{l\in \ell_{rec}}$, and $\{\lambda_l\}_{l\in \ell_{tex}}$ are set to be 1 for all layers.

\textbf{Evaluation protocol.} We follow the standard protocols for unsupervised domain adaptation as \cite{Ganin2015Unsupervised,Long2015Learning}. All labeled source characters and all unlabeled target characters are utilized for training, and the average classification accuracy and standard deviation are reported based on three random experiments.

\subsection{Ablation study}

To take a further step into different components within our method, we report the results of 5 variations of STSN in Table \ref{ablation}. Specifically, we remove some of $\mathcal{L}_{advF}$, $\mathcal{L}_{advI}$, $\mathcal{L}_{rec}$, $\mathcal{L}_{per}$ and $\mathcal{L}_{cls}^{st}$ from the full model. From the results, we can find that each component makes an important contribution to the target performance and the quality of transformed images.

\textbf{Effectiveness of alignment (Model-A v.s. STSN).} We first investigate the effectiveness of $\mathcal{L}_{advF}$ which aligns source and target features in structure-related space. We perform ablation study with Model-A, which is built by only eliminating $\mathcal{L}_{advF}$. Compared with STSN, the performance of Model-A drops 15.3\%. It shows that alignment between the two domains indeed helps to improve the model generalization. Moreover, $\mathcal{L}_{advF}$ makes structure features domain-invariant and texture-independent leading to better disentanglement.

\textbf{Effectiveness of disentanglement (Model-B v.s. STSN).} Model-B removes all feature disentanglement modules from STSN, and thus it only globally aligns source and target domains in the whole feature space. From the comparison results between Model-B and our STSN, we can see that dramatic decrease happens in the target accuracy of Model-B. This result reveals the effect of disentanglement is important for adaptation process, since alignment can be adopted in structure-related feature space so that adaptation process is protected from being contaminating by texture leading to the performance improvement.

\textbf{Effectiveness of transformation (Model-E v.s. STSN).} STSN further optimizes the network using the transformed target-like images supervised by $\mathcal{L}_{cls}^{st}$. To explore its effect, we also perform ablation study with Model-E, i.e., STSN w/o $\mathcal{L}_{cls}^{st}$. Compared with STSN, Mode-E suffers a degradation of 4.4\% w.r.t. target accuracy, manifesting the importance of the proposed transformation module. The transformed target-like images achieve data augmentation and label transfer, and supervision on them helps to reduce intra-class variance in target domain.

\begin{figure}
\centering
\subfigure[Source]{
\label{ablation_transform_a} 
\includegraphics[width=1.5cm]{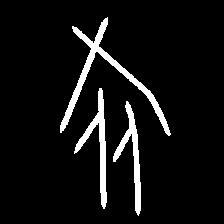}}
\hspace{0.2cm}
\subfigure[Model-C]{
\label{ablation_transform_b} 
\includegraphics[width=1.5cm]{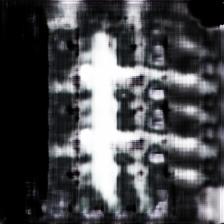}}
\hspace{0.2cm}
\subfigure[Model-D]{
\label{ablation_transform_c} 
\includegraphics[width=1.5cm]{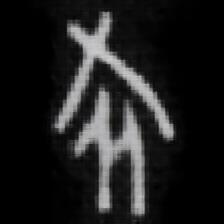}}
\hspace{0.2cm}
\subfigure[Model-E]{
\label{ablation_transform_d} 
\includegraphics[width=1.5cm]{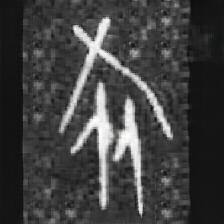}}
\caption{Ablation: the effect of $\mathcal{L}_{advI}$, $\mathcal{L}_{rec}$ and $\mathcal{L}_{per}$ on transformation. (a) Source handprinted character. (b) Without $\mathcal{L}_{rec}+\mathcal{L}_{per}$, the transformed image is not satisfied. (c) Without $\mathcal{L}_{per}$, the transformed image lacks the proper texture. (d) With all the constraints, the transformed image looks real from both structure and texture aspects.}
\label{ablation_transform} 
\end{figure}

\textbf{Quality of transformed image (Model-C v.s. Model-D v.s. Model-E).} We further analyze the effect of $\mathcal{L}_{advI}$, $\mathcal{L}_{rec}$ and $\mathcal{L}_{per}$ on transformation. Model-C represents the transformation module w/o $\mathcal{L}_{rec}+\mathcal{L}_{per}$, Model-D denotes the module w/o $\mathcal{L}_{per}$, and Mode-E is the full transformation module. The transformed images generated by these models are visualized in Fig. \ref{ablation_transform}. Fig. \ref{ablation_transform_b} demonstrates Model-C with $\mathcal{L}_{advI}$ alone fails to produce the desired behavior for our task. When adding $\mathcal{L}_{rec}$, Model-D can generate images with the proper glyph information but still fails to match the target style due to the absence of a proper constraint of texture similarity. In Fig. \ref{ablation_transform_d}, the full transformation module can generate a satisfactory transformed image, indicating that $\mathcal{L}_{advI}$, $\mathcal{L}_{rec}$ and $\mathcal{L}_{per}$ all make important contributions to transformation. Moreover, we assume that the quality of transformed images will affect feature disentanglement and so the adaptation performance. Therefore, we also report the corresponding adaptation results of Model-C, Model-D and Model-E in Table \ref{ablation}. It can be observed Model-E achieves the best performance among them which also proves our assumption. 

\begin{figure}
\centering
\subfigure[Sensitivity to $\alpha_1$]{
\label{sensi_a} 
\includegraphics[width=4cm]{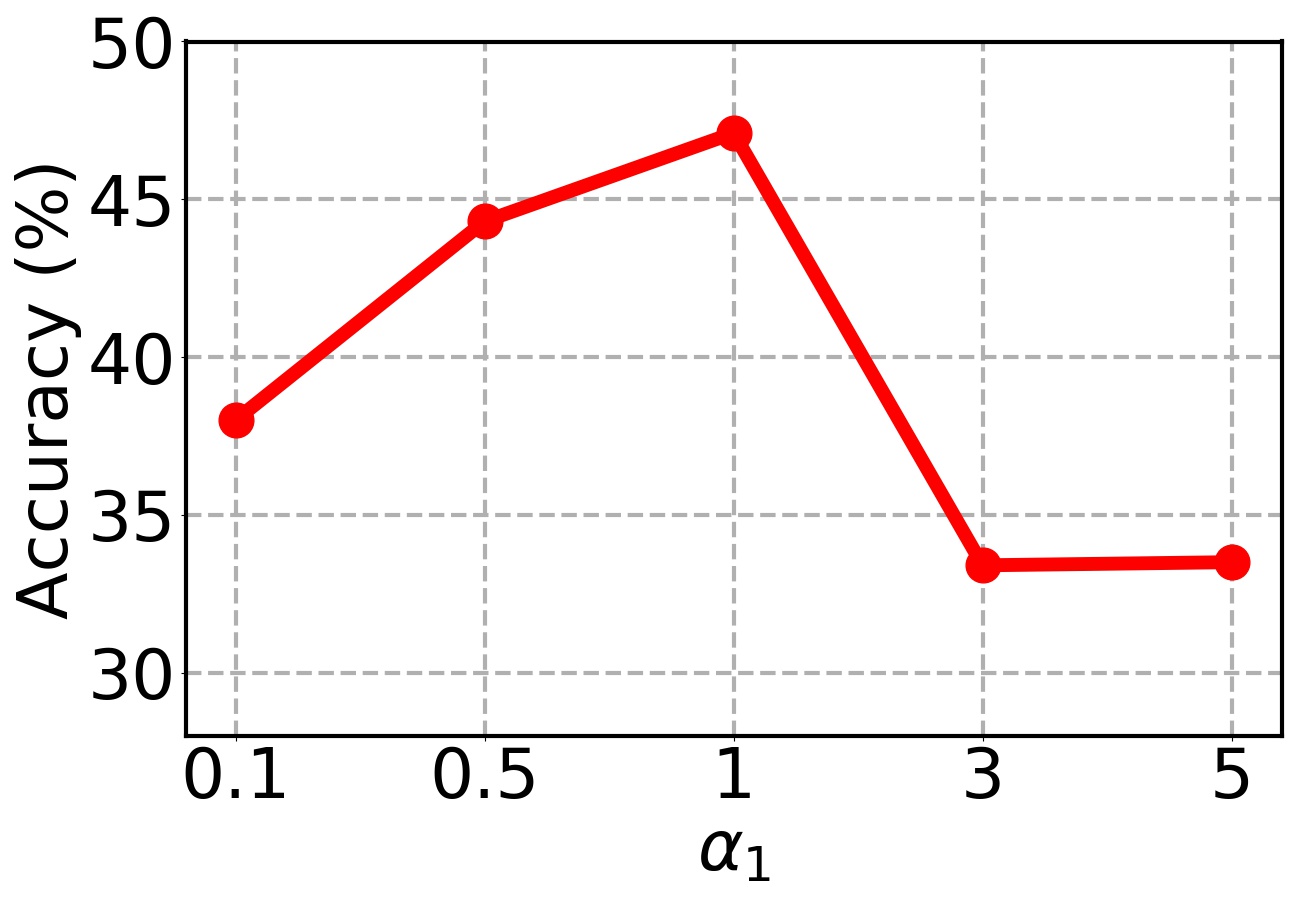}}
\hspace{0cm}
\subfigure[Sensitivity to $\alpha_2$]{
\label{sensi_b} 
\includegraphics[width=4cm]{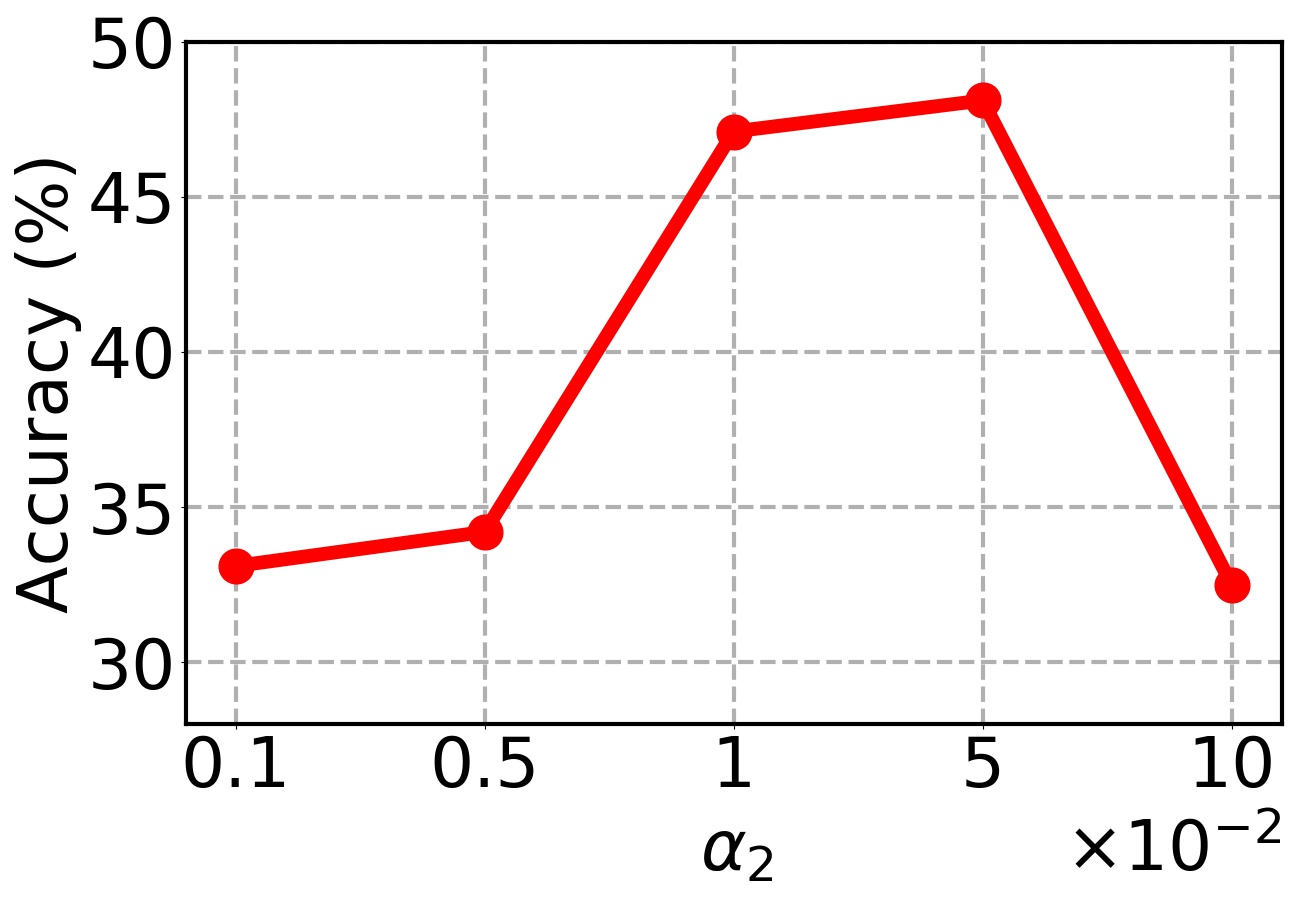}}
\hspace{0cm}
\subfigure[Sensitivity to $\alpha_3$]{
\label{sensi_b} 
\includegraphics[width=4cm]{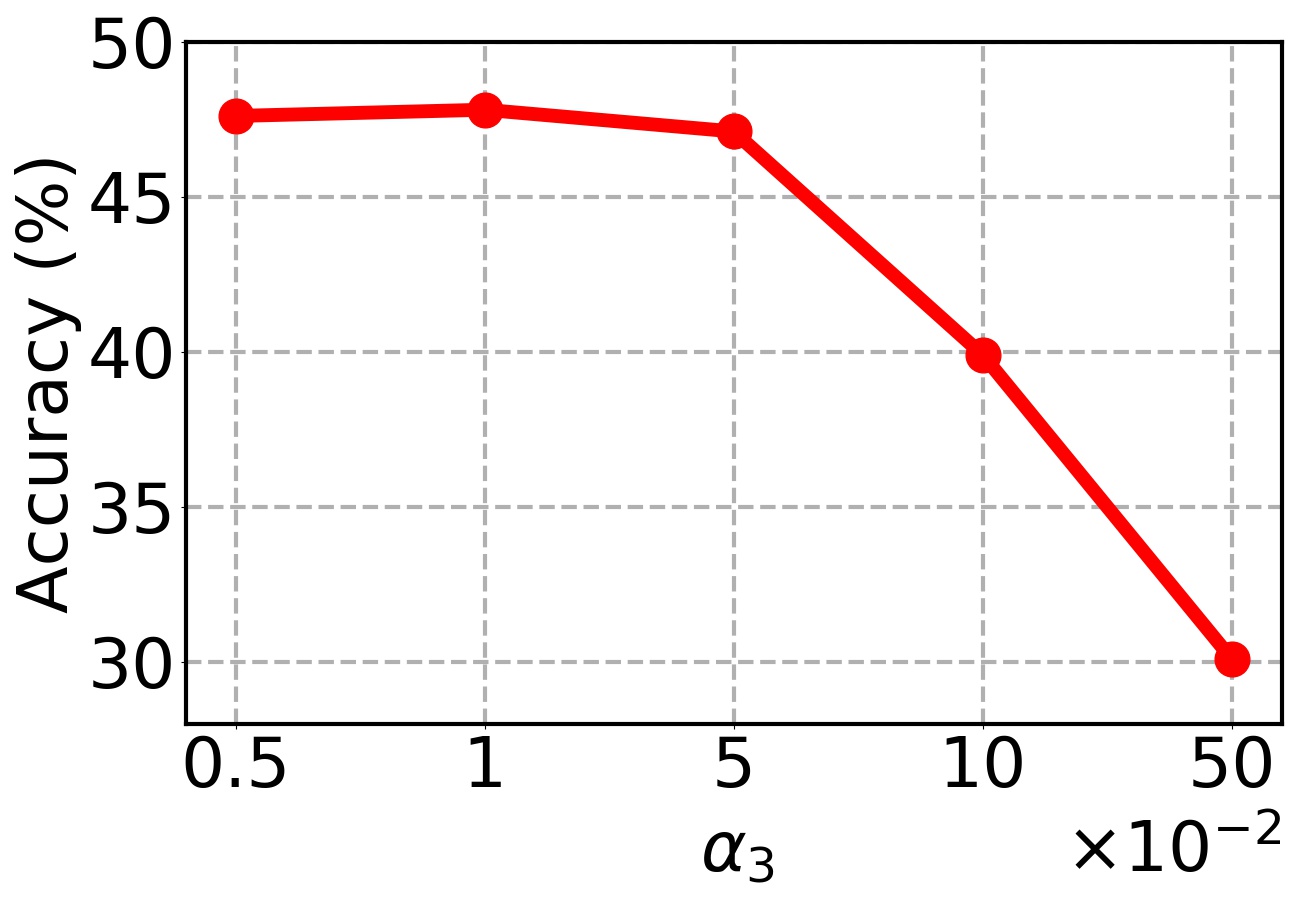}}
\hspace{0cm}
\subfigure[Sensitivity to $\alpha_4$]{
\label{sensi_b} 
\includegraphics[width=4cm]{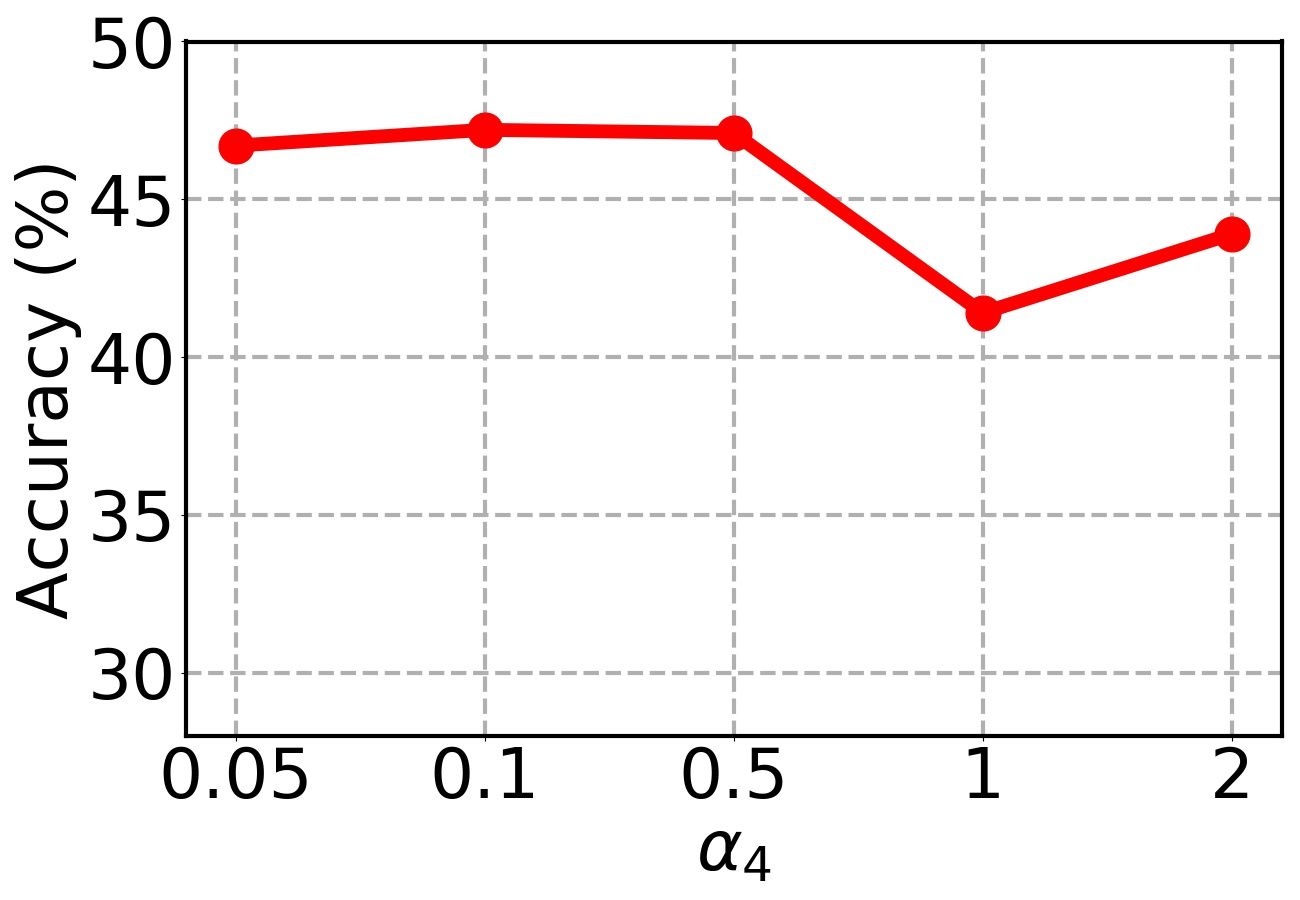}}
\caption{The sensitivity of target accuracy to $\{\alpha_i\}_{i=1}^4$ when adapting handprinted oracle data to scanned data.}
\label{sensi} 
\end{figure}

\textbf{Parameter Sensitivity.} $\{\alpha_i\}_{i=1}^4$ are the trade-off parameters for different losses in Eq. (\ref{total}). We investigate the target performance of STSN when these parameters change, and show the results in Fig. \ref{sensi}. It can be observed that the accuracy first increases and then decreases as $\alpha_1$ and $\alpha_2$ vary, demonstrating a desirable bell-shaped curve. In principle, smaller value of $\alpha$ would avoid STSN learning the disentangled and aligned features well; while larger value of $\alpha$ will overemphasize on alignment and lead to negative transfer. We also observe that a small $\alpha_3$ (e.g., 0.05) tends to have a better performance, but the accuracy would suffer an obvious decline when $\alpha_3>0.05$, since a larger $\alpha_3$ will weaken the effect of $\mathcal{L}_{cls}^{s}$ term and degrade the classification performance. For $\alpha_4$, the accuracy is stable under a wide range of parameter values, indicating STSN is not quite sensitive to $\alpha_4$.

\begin{table}
\small
\renewcommand\arraystretch{1.1}
\caption{The empirical analysis results for the sensitivity of hyperparameter $\lambda_l$}
\label{lambda_sensi}
	\begin{center}
    \small
    \setlength{\tabcolsep}{4mm}{
	\begin{tabular}{l|c|ccc}
		\hline
         Methods & Domain &  ASC  & SAME & DES  \\ \hline \hline
         \multirow{2}{*}{$\{\lambda_l\}_{l\in \ell_{rec}}$} &Source & \textbf{95.0} & 94.2 & 89.6 \\
         & Target & \textbf{47.1} & 43.7 & 43.1 \\ \hline
         \multirow{2}{*}{$\{\lambda_l\}_{l\in \ell_{str}}$} &Source & \textbf{95.0} & 94.1 & 94.1 \\
         & Target & \textbf{47.1} & 32.0 & 36.3 \\ \hline
         \multirow{2}{*}{$\{\lambda_l\}_{l\in \ell_{tex}}$} &Source & 94.2 & \textbf{95.0} & 94.6 \\
         & Target & 34.3 & \textbf{47.1} & 35.0 \\
         \hline \hline
	\end{tabular}}
    \end{center}
\end{table}

Moreover, we also check the sensitivities of $\lambda_l$ in $\mathcal{L}_{rec}$ and $\mathcal{L}_{per}$. In our method, the values of $\{\lambda_l\}_{l\in \ell_{rec}}$ are increased along with the layer number, and we call it ASC model. We build two variants, i.e., SAME and DEC, by changing the values of $\lambda_l$, and compare ASC with them shown in Table \ref{lambda_sensi}. $\{\lambda_l\}_{l\in \ell_{rec}}$ keep the same for all layers in SAME model, but are decreased along with the layer number in DEC model. From the results, we can see that ASC model outperforms other comparison models. Since the preservation of content and semantic information is more difficult but meaningful for oracle recognition, we prefer to pay more attention to higher layers in our feature reconstruction loss. Similar experiments are performed on $\{\lambda_l\}_{l\in \ell_{str}}$ and $\{\lambda_l\}_{l\in \ell_{tex}}$. We observe that ASC model also performs best in structure loss, while SAME model is superior to others in texture loss. Therefore, it is necessary to set proper weight for each layer.

\subsection{Visualization}

\begin{figure}
\centering
\subfigure[Source domain]{
\label{Convergence_a} 
\includegraphics[width=4cm]{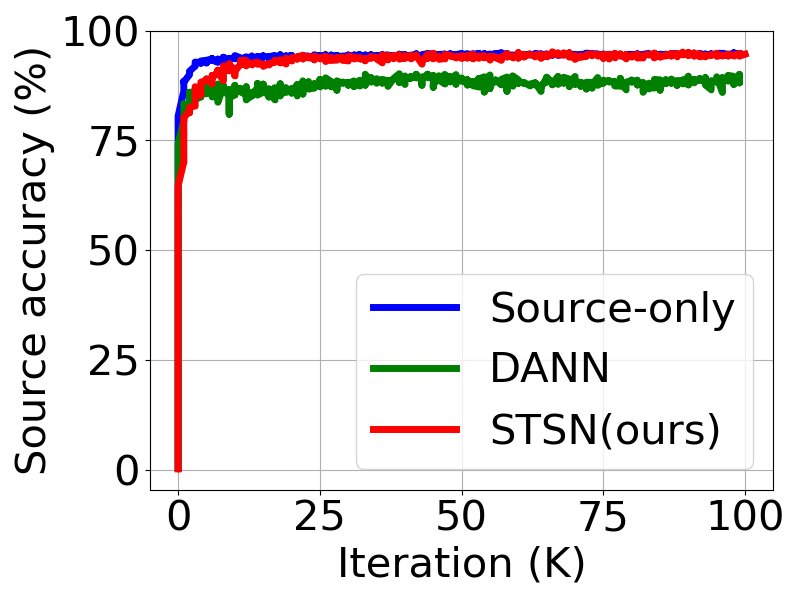}}
\hspace{0cm}
\subfigure[Target domain]{
\label{Convergence_b} 
\includegraphics[width=4cm]{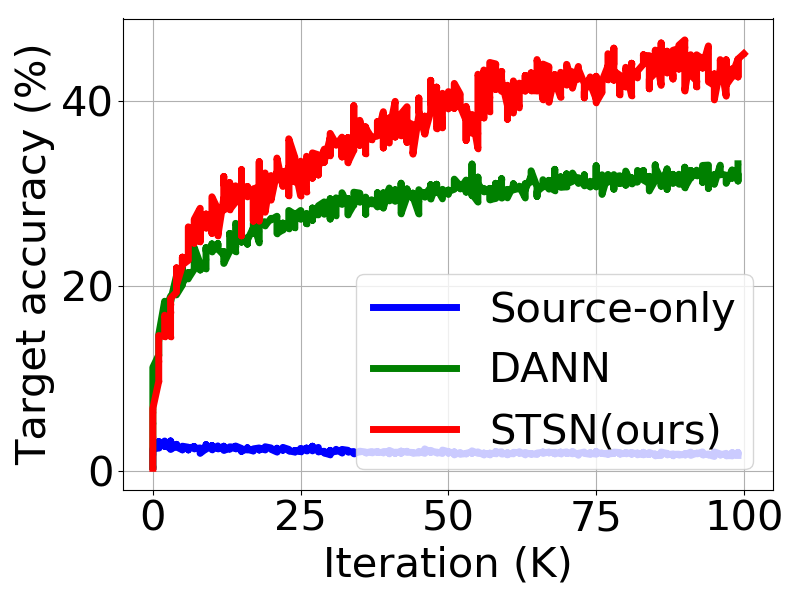}}
\caption{The testing accuracies of Source-only, DANN and our proposed method evaluated on source and target domain when adapting handprinted oracle characters to scanned data.}
\label{Convergence} 
\end{figure}

\begin{figure*}
\centering
\includegraphics[width=16cm]{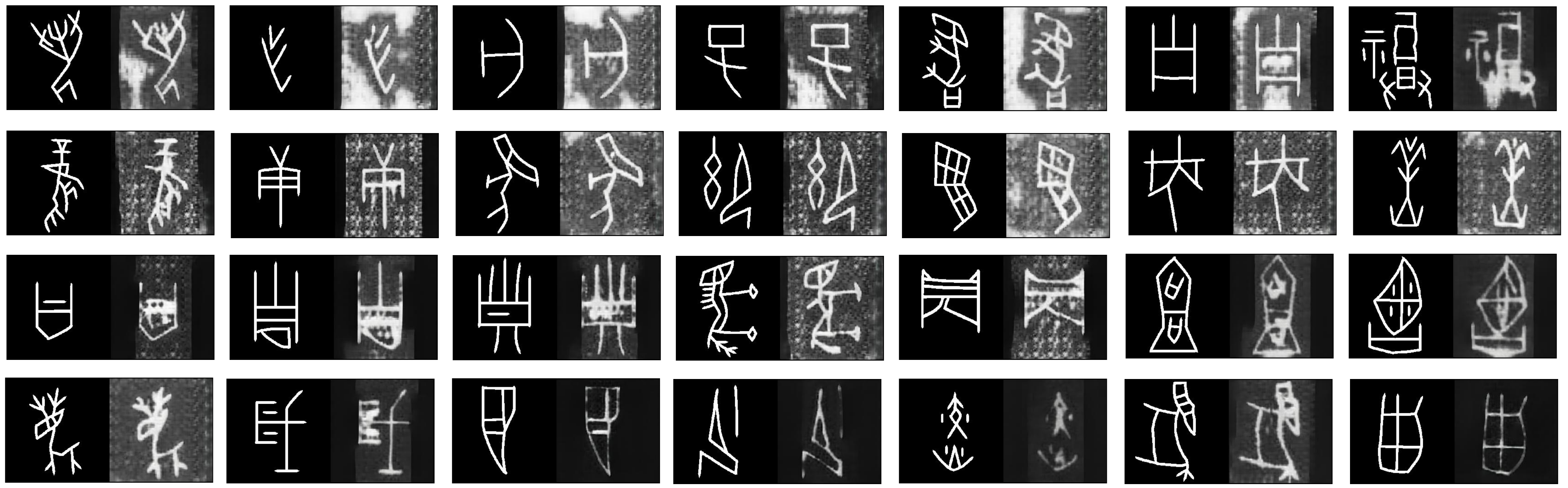}
\caption{The transformed target-like images on Oracle-241 dataset. In every two columns, the left and right images are the images from real handprinted domain and their corresponding transformed images in target domain.}
\label{transfromed_scan} 
\end{figure*}

\begin{figure*}
\centering
\includegraphics[width=16cm]{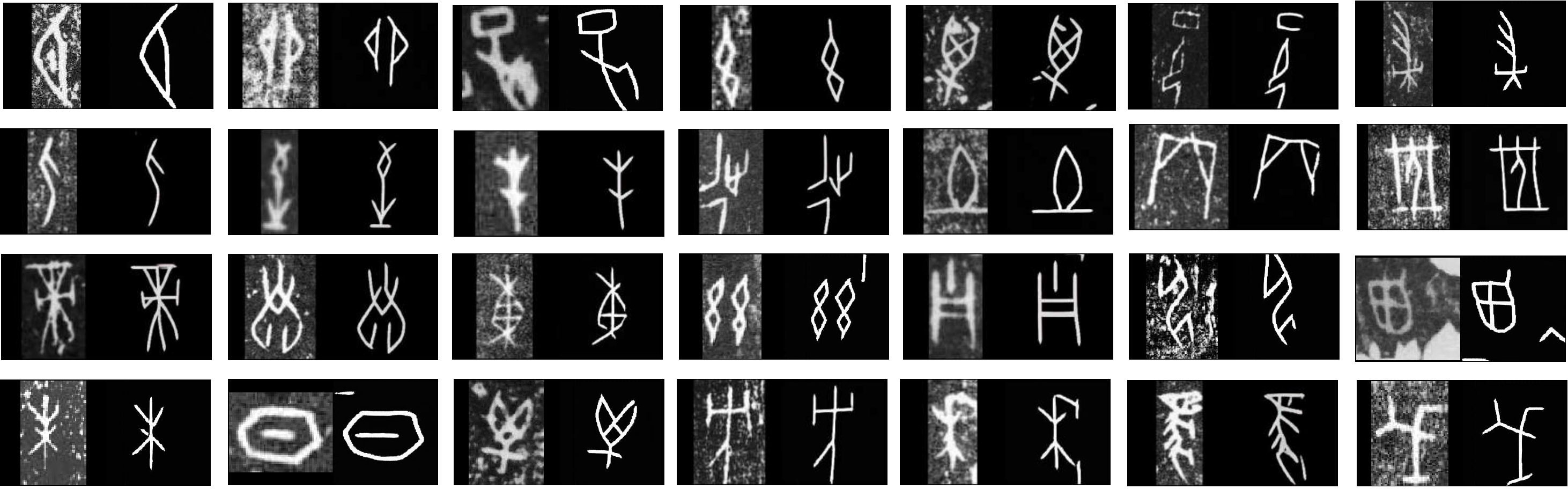}
\caption{The transformed source-like images on Oracle-241 dataset. In every two columns, the left and right images are the images from real scanned domain and their corresponding transformed images in source domain.}
\label{transfromed_handprint} 
\end{figure*}

\begin{figure}
\centering
\includegraphics[width=8.5cm]{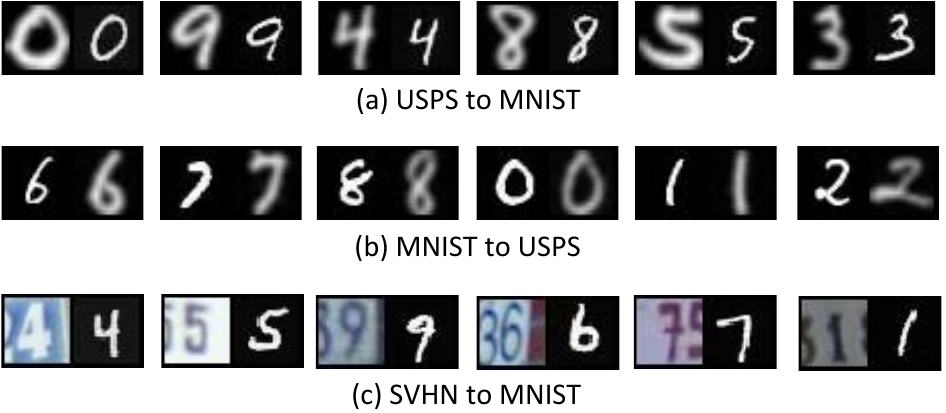}
\caption{The transformed target-like images on (a) U$\rightarrow $M, (b) M$\rightarrow $U and (c) S$\rightarrow $M tasks. In every two columns, the left and right images are the images from source domain and their corresponding transformed images in target domain.}
\label{transform_digit} 
\end{figure}

\begin{figure}[h]
\centering
\subfigure[Before adaptation]{
\label{oracle_visual_a} 
\includegraphics[width=4cm]{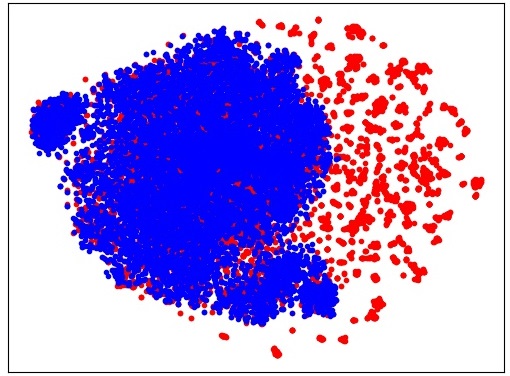}}
\hspace{0cm}
\subfigure[After adaptation]{
\label{oracle_visual_b} 
\includegraphics[width=4cm]{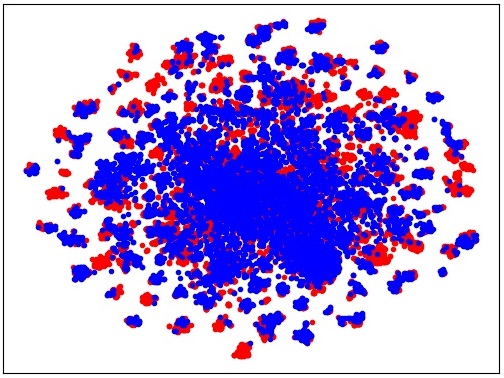}}
\caption{Feature visualization on testing data of Oracle-241 dataset. Red points are handprinted samples and the blue ones represent scanned samples. (Best viewed in color)}
\label{oracle_visual} 
\end{figure}

\begin{figure}[h]
\centering
\subfigure[Before adaptation]{
\label{digit_visual_a} 
\includegraphics[width=4cm]{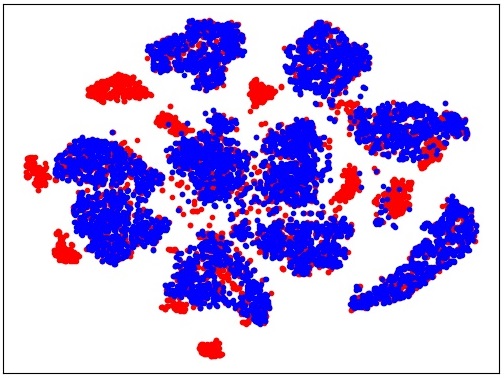}}
\hspace{0cm}
\subfigure[After adaptation]{
\label{digit_visual_b} 
\includegraphics[width=4cm]{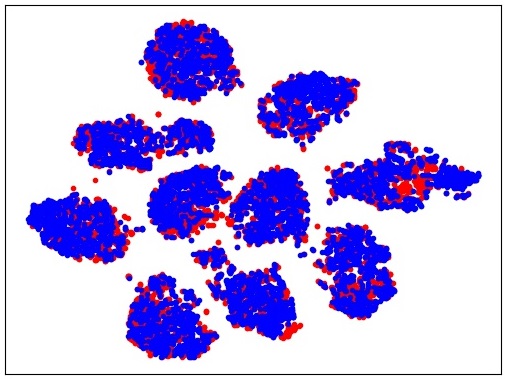}}
\caption{Feature visualization on U$\rightarrow $M task. Red points are USPS samples and the blue ones represent MNIST samples. (Best viewed in color)}
\label{digit_visual} 
\end{figure}

\begin{figure}
\centering
\subfigure[Before adaptation]{
\label{intra_visual_a} 
\includegraphics[width=4cm]{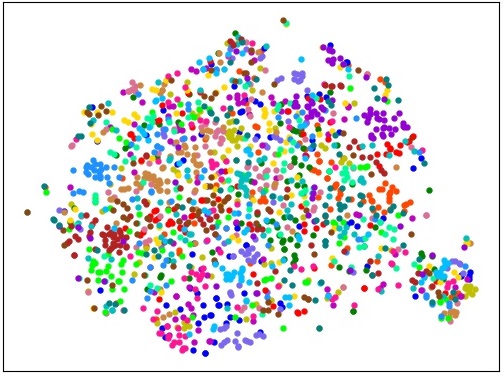}}
\hspace{0cm}
\subfigure[After adaptation]{
\label{intra_visual_b} 
\includegraphics[width=4cm]{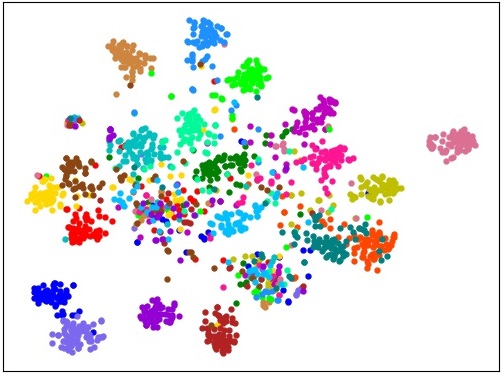}}
\caption{Feature visualization on target scanned data of Oracle-241 dataset. The color of a circle represents its class. 
(Best viewed in color)}
\label{intra_visual} 
\end{figure}

\textbf{Convergence.} We empirically study the convergence property of the proposed model. Fig. \ref{Convergence} illustrates the corresponding test accuracies with respect to the number of iterations. From the figure, we can see that the accuracy is monotonically increasing with the increase of iterations, and STSN gets converged after several iterations. STSN achieves much better results than DANN \cite{Ganin2015Unsupervised}, but shows lower convergence rate on target domain. We conjecture the reason is as follows. Our STSN boosts target performance by disentanglement, alignment and transformation. However, both disentanglement and transformation largely depend on the generator $G$. 
Absolute disentanglement and realistic transformed images cannot even be achieved until $G$ converges. Therefore, the convergence rate of STSN may be slowed down due to the optimization of the generator $G$.

\textbf{Image visualization.} Fig. \ref{transfromed_scan} visualizes the transformed target-like images of our method on Oracle-241. 
As seen, our method can effectively generate the target-like images with different types of noises and abrasions. The images in the first row are transformed into some incomplete characters covered by some white regions. The second row shows the transformed images suffered from serious abrasions similar to salt and pepper noise. In the third row, the generated images are contaminated by the added strokes or the omitted strokes. Moreover, scanned characters are generated by placing a piece of paper over oracle-bone and then rubbing the paper with rolled ink. The manually applied pressure while rubbing causes the variations of contrast and makes strokes differ in thickness. This type of transformed images is shown in the last row. 
Fig. \ref{transfromed_handprint} shows some transformed source-like images which also proves the effectiveness of our transformation method. Although these denoising images are not utilized to optimize network, they play important roles in disentanglement. We also show the transformed images on MNIST-USPS-SVHN datasets in Fig. \ref{transform_digit}. These transformed images all look alike the real target images regarding texture while successfully retaining the same glyph information as source images.

\textbf{Feature visualization.} To demonstrate the effectiveness of our method on reducing domain shift, we extract source and target features of Oracle-241 dataset using ``Source-only" and our STSN model, and visualize the learned features using t-SNE \cite{maaten2008visualizing}. ``Source-only" model is trained on handprinted data without any adaptation. The results are shown in Fig. \ref{oracle_visual}. As we can see, the distributions are separated between domains before adaptation. 
After adaptation, features are successfully fused and thus there is no domain discrepancy between different domains. The similar observation can be obtained on digit datasets shown in Fig. \ref{digit_visual}. Furthermore, to demonstrate the effectiveness of our method on reducing intra-class variance, we also visualize target features on Oracle-241 using t-SNE. We randomly select some scanned oracle characters of 20 classes, then extract and visualize their features by ``Source-only" and our STSN model. The results are shown in Fig. \ref{intra_visual}. We can observe that features belonging to different classes are mixed together before adaptation while our STSN features are more discriminative. The features in the same class are mapped close and features of different classes are separated to some extent.

\subsection{Comparison with state-of-the-arts}

\begin{figure*}
\centering
\includegraphics[width=17cm]{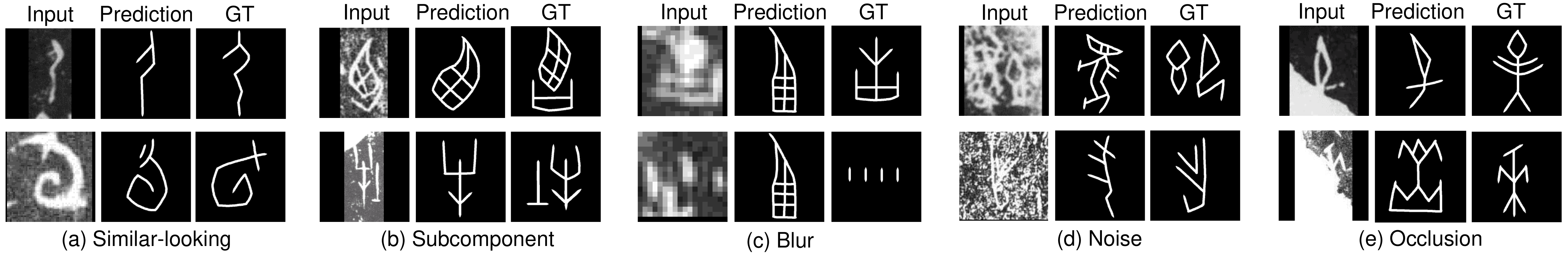}
\caption{Ten example samples which are misclassified by our STSN model. For each character, the left, middle and right images are the misclassified scanned sample, prediction and ground-truth, respectively. The predictions and ground-truths are showed visually by the corresponding handprinted images belonging to the same classes.}
\label{error_img} 
\end{figure*}

\begin{figure}
\centering
\includegraphics[width=7.8cm]{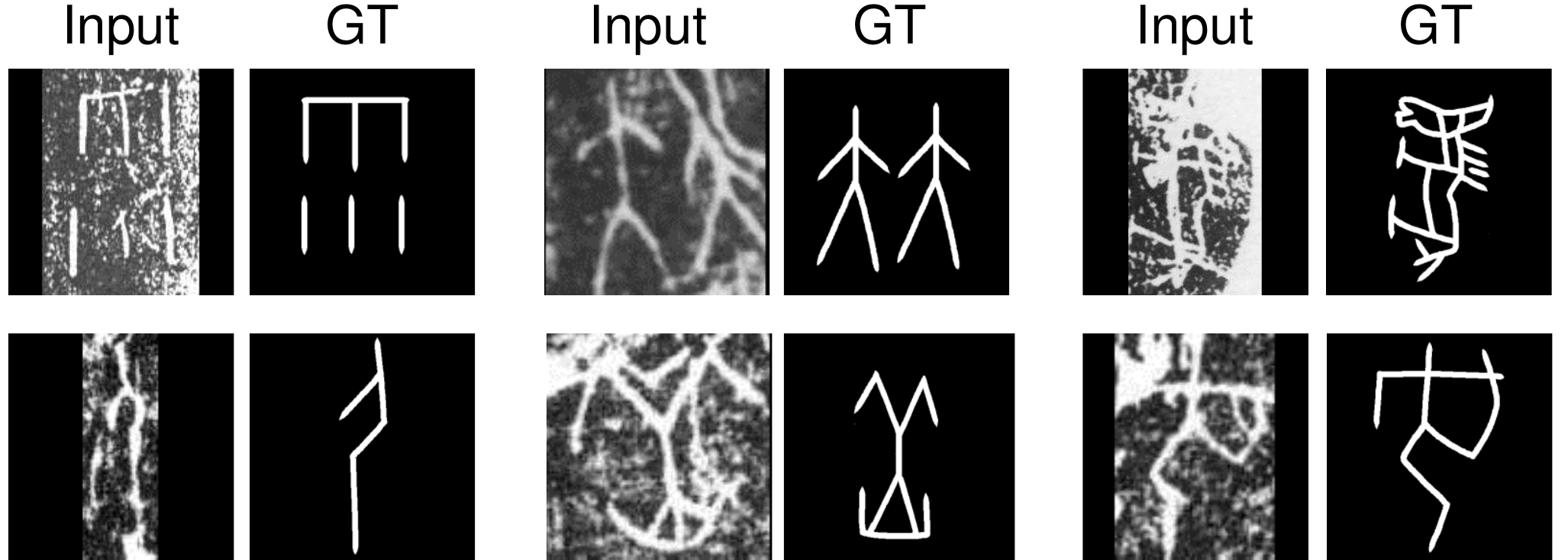}
\caption{Six example images which are misclassified by ``Source-only" model but classified correctly by our method. The ground-truths are showed by the corresponding handprinted images belonging to the same classes.}
\label{correct_img} 
\end{figure}

\textbf{Results on Oracle-241.} Here we transfer recognition knowledge from handprinted oracle characters to scanned data. To prove its effectiveness, our STSN is compared with some methods designed for oracle character recognition \cite{huang2019obc306,zhang2019oracle}. Since they only trained models on handprinted data without any adaptation, we call them ``Source-only" models. Several SOTA adaptation methods used for image classification, e.g., CDAN \cite{long2018conditional}, BSP \cite{chen2019transferability} and GVB \cite{cui2020gradually}, are also compared. 
All of them are utilized ResNet-18 as backbone for fair comparison. The quantitative results are described as follows.

\begin{table}[htbp]
\renewcommand\arraystretch{1.1}
\caption{Source and target accuracy (mean$\pm$std\%) on Oracle-241 dataset (the handprinted characters to scanned data setting). The best numbers are indicated in bold.}
 \label{oracle-241}
	\begin{center}
    \small
    \setlength{\tabcolsep}{3.2mm}{
	\begin{tabular}{c|c|ccc}
		\hline
          \multicolumn{2}{c|}{Methods} & Handprint & Scan  \\ \hline \hline
         \multirow{2}{*}{Source-only} & ResNet \cite{huang2019obc306,he2016deep} & 94.9$\pm $0.1  &  2.1$\pm $0.6   \\
         &Zhang et al. \cite{zhang2019oracle} & 94.5$\pm $0.4  &  8.4$\pm $1.0   \\\hline
         \multirow{7}{*}{Adaptation} & CORAL \cite{sun2016deep} & 89.5$\pm $0.6 &  18.4$\pm $1.3 \\
         & DDC \cite{tzeng2014deep} & 90.8$\pm $1.5  &   25.6$\pm $1.9   \\
         & DAN \cite{Long2015Learning} & 90.2$\pm $1.5  &   28.9$\pm $1.6   \\
         & ASSDA \cite{zhang2021robust} &85.8$\pm $0.1 &32.6$\pm $0.2 \\
         & DANN \cite{Ganin2015Unsupervised} & 87.1$\pm $1.7 &  32.7$\pm $1.5     \\
         & GVB \cite{cui2020gradually} & 92.8$\pm $0.4  & 36.8$\pm $1.1  \\
         & CDAN \cite{long2018conditional} & 85.3$\pm $3.3 & 37.9$\pm $2.0  \\
         & BSP \cite{chen2019transferability} &87.7$\pm $0.7 & 43.7$\pm $0.4  \\ \cline{2-4}
         & \textbf{STSN (ours)} & \textbf{95.0$\pm $0.2} &  \textbf{47.1$\pm$0.8}   \\ \hline
	\end{tabular}}
    \end{center}
\end{table}

\begin{table}
\renewcommand\arraystretch{1.1}
\caption{Target accuracy (mean$\pm$ std\%) on three transfer tasks of digit datasets. The best numbers are indicated in bold.}
\label{MNIST-USPS-SVHN}
	\begin{center}
    \small
    \setlength{\tabcolsep}{2mm}{
	\begin{tabular}{l|ccc|c}
		\hline
         Methods &   U$\rightarrow $M  & M$\rightarrow $U & S$\rightarrow $M & Avg \\ \hline \hline
         Source-only \cite{he2016deep} & 69.6$\pm$3.8 & 82.2$\pm$0.8 & 67.1$\pm$0.6 & 73.0 \\
         DSN w/ MMD \cite{bousmalis2016domain} & - & - & 72.2& - \\
         DANN \cite{Ganin2015Unsupervised} & - & 77.1$\pm$1.8 & 73.6& - \\
         ADDA \cite{Tzeng2017Adversarial} & 90.1$\pm$0.8 & 89.4$\pm$0.2 & 76.0$\pm$1.8 & 85.2 \\
         DAA \cite{jia2019domain} &  92.8$\pm$1.1 & 90.3$\pm$0.2 & 78.3$\pm$0.5 & 87.1 \\
         LEL \cite{luo2017label} & - & - & 81.0$\pm$0.3& - \\
         DSN w/ DANN \cite{bousmalis2016domain} & - & - & 82.7& - \\
         DTN \cite{taigman2016unsupervised} & - & - & 84.4& - \\
         AsmTri \cite{saito2017asymmetric} &   - & - & 86.0& - \\
         CoGAN \cite{liu2016coupled}  & - & - & 91.2$\pm$0.8& - \\
         MSTN \cite{xie2018learning} & - & 92.9$\pm$1.1 & 91.7$\pm$1.5& - \\
         PixelDA \cite{bousmalis2017unsupervised} & & \textbf{95.9} & & -\\
         UNIT \cite{liu2017unsupervised} & 93.6 & \textbf{95.9} & 90.5 & 93.4\\
         CyCADA \cite{hoffman2018cycada} & 96.5$\pm$0.1 & 95.6$\pm$0.2 & 90.4$\pm$0.4 & 94.2 \\
         \textbf{STSN (ours)} &  \textbf{96.7$\pm$0.1} &  94.4$\pm$0.3 &  \textbf{92.2$\pm$0.1} & \textbf{94.4}  \\
         \hline \hline
	\end{tabular}}
    \end{center}
\end{table}

From the results shown in Table \ref{oracle-241}, we have the following observations. Firstly, when trained on handprinted data and tested on the same domain, ``Source-only" models can achieve high accuracies. However, the performances dramatically decrease when they are directly applied on scanned data, which brings large challenges on recognizing real-world oracle characters. 
Although Zhang et al. \cite{zhang2019oracle} introduced triplet loss and nearest neighbor classifier to oracle character recognition, they ignored the distribution discrepancy between handprinted and scanned data. STSN is the first one to discuss this issue, and applies UDA in oracle character recognition.

Secondly, we notice that existing UDA methods can successfully improve target performance by learning a domain-invariant feature space. 
This phenomenon shows the importance of mitigating the domain shift. However, this improvement is still limited for target domain and the source accuracy is decreased unfortunately after adapting. For example, ASSDA \cite{zhang2021robust} which adopted both image-level and character-level alignment only achieves 32.6\% on target domain. Without consideration of texture-specific information contained in each domain, it is quite challenging to align the whole source and target domains well. Moreover, hard examples with excessive domain characteristics would have negative effect in alignment and so the performances of both two domains. Although GVB \cite{cui2020gradually} published in CVPR'20 utilized a bridge, i.e., one fully connected layer, to model domain-specific parts, the bridge with simple architecture is hard to capture these properties well.

Lastly, benefitting from joint adaptation, disentanglement and transformation, our STSN outperforms other competitors significantly, which proves its effectiveness. To be specific, it achieves 47.1\%  when tested on scanned data, which is higher than CDAN \cite{long2018conditional} by 9.2\% and higher than the best baseline BSP \cite{chen2019transferability} by 3.4\%. Moreover, it also maintains better performance on source domain after adapting because alignment in structure-related feature space protects adaptation process from the contamination by texture.

\textbf{Results on MNIST-USPS-SVHN.} 
We additionally perform UDA experiments on other text datasets, i.e., MNIST-USPS-SVHN. We follow the evaluation protocol in \cite{hoffman2018cycada,long2018conditional} and report the target accuracy on each transfer task. Results of digit datasets are shown in Table \ref{MNIST-USPS-SVHN}. DSN \cite{bousmalis2016domain} which disentangled features by image reconstruction and feature orthogonality only achieves 82.7\% on S$\rightarrow$M task. 
CyCADA \cite{hoffman2018cycada} which took advantage of CycleGAN \cite{zhu2017unpaired} to generate the transformed target-like images largely boosts the target performance and achieves 90.4\% on S$\rightarrow$M task. Our proposed model outperforms all comparison methods on almost all the tasks. The utilization of transformation to constrain the disentanglement process enables the absolute separation and the safe alignment. Instead of using the binary domain labels, transforming images using the learned texture information makes our transformed images more realistic and diverse such that intra-class variances can be better reduced. After adaptation, our method achieves an accuracy of 96.7\% on U$\rightarrow $M task, and achieves 92.2\% on S$\rightarrow $M task. The results further demonstrate the advantage of our STSN method.

\subsection{Error analysis}

Some example images which are misclassified by ``Source-only" model \cite{he2016deep} but classified correctly by our STSN are presented in Fig. \ref{correct_img}. From the results, we can find that ``Source-only" model fails to recognize the scanned images with serious noises and abrasions while our STSN can succeed. However, more than 50\% of scanned images are still misclassified by our method. We show some of them in Fig. \ref{error_img}, and conjecture the reasons may be as follows. First, our model is sometimes confused by some similar-looking characters. For example, the characters of the predicted category and the ground-truth (GT) category only differ in some small details shown in Fig. \ref{error_img} (a). Second, some characters are subcomponents of the others which also makes our model more error prone as shown in Fig. \ref{error_img} (b). Third, the images are seriously degraded and even completely lost their discriminative glyph information caused by blur (Fig. \ref{error_img} (c)), noise (Fig. \ref{error_img} (d)) and occlusion (Fig. \ref{error_img} (e)). Recognizing these characters is challenging even for human.

\section{Conclusion and future work}

To address the problem of lack of massive labeled real-world scanned oracle data, we are the first to apply unsupervised domain adaptation in oracle character recognition. The main contribution of this paper is that we present a publicly available oracle character database for domain adaptation, and propose a structure-texture separation network (STSN) to transfer recognition knowledge from handprinted oracle characters to real scanned data. STSN is an end-to-end learning framework for joint disentanglement, transformation, adaptation and recognition. Unlike previous approaches which globally align the two domains, our STSN disentangles features and applies alignment in structure-related space. Moreover, domain transformation by swapping textures is adopted to achieve data augmentation, reduce intra-class variance and guarantee the absolute separation of features. 
Experiments on our proposed Oracle-241 dataset have convincingly demonstrated that STSN successfully improves the performance on scanned data and outperforms several SOTA methods. 

However, the accuracy on scanned data is less than 50\%, and there is still room for improvement. Hence, one future trend is to investigate more effective and efficient methods, e.g., pseudo-labels, to further improve the generalization ability of recognition model. Moreover, considering that it is possible to access a small amount of labeled scanned data, we aim to extend our method from the unsupervised setting to semi-supervised setting in the future.

Recognizing oracle data accurately is an open issue for both human and automatic system. Human expert work is still needed, but our system can lighten their workload. Given an oracle character, our STSN can provide the predicted probability to experts and help them with decision-making. When the system confidently identifies which category the sample belongs to, experts only need to check the result. When the system fails to give a confident prediction, this sample may be difficult to recognize (or has not been deciphered) and thus needs to be further identified by experts. Considering that about 2,200 characters have been successfully deciphered, we plan to cooperate with some experts and collect more oracle characters to cover all of the 2,200 categories such that the trained system can be more practical. 

\section{Acknowledgments}

This work was partially supported by the National Natural Science Foundation of China under Grants No. 61871052 and 62192784.

{
\bibliographystyle{IEEEtran}
\bibliography{egbib}
}

\end{document}